r
\documentclass{article}

\usepackage{microtype}
\usepackage{graphicx}
\usepackage{booktabs} 

\usepackage{hyperref}

\usepackage[accepted]{icml2020}

\usepackage[utf8]{inputenc} 
\usepackage[T1]{fontenc}    
\usepackage{hyperref}       
\usepackage{url}            
\usepackage{booktabs}       
\usepackage{amsfonts}       
\usepackage{nicefrac}       
\usepackage{microtype}      
\usepackage{amssymb}
\RequirePackage{algorithm}
\RequirePackage{algorithmic}

\RequirePackage{fancyhdr}
\RequirePackage{color}
\RequirePackage{algorithm}
\RequirePackage{algorithmic}
\RequirePackage{natbib}
\RequirePackage{eso-pic} 
\RequirePackage{forloop}
\usepackage{microtype}
\usepackage{graphicx}
\usepackage{tabularx}
\usepackage{subcaption}
\usepackage{booktabs} 
\usepackage{amsmath}
\usepackage[normalem]{ulem}

\usepackage{color}
\definecolor{red}{RGB}{190, 59, 55}

\usepackage{hyperref}

\icmltitlerunning{Prioritized Sequence Experience Replay}

\begin{document}

\twocolumn[
\icmltitle{Prioritized Sequence Experience Replay}
\icmlsetsymbol{equal}{*}

\begin{icmlauthorlist}
\icmlauthor{Marc Brittain}{equal,to}
\icmlauthor{Josh Bertram}{equal,to}
\icmlauthor{Xuxi Yang}{equal,to}
\icmlauthor{Peng Wei}{to}
\end{icmlauthorlist}

\icmlaffiliation{to}{Department of Aerospace Engineering, Iowa State University, Ames, USA}

\icmlcorrespondingauthor{Marc Brittain}{mwb@iastate.edu}

\icmlkeywords{Deep Reinforcement Learning, Experience Replay}

\vskip 0.3in
]



\printAffiliationsAndNotice{\icmlEqualContribution} 

\begin{abstract}
Experience replay is widely used in deep reinforcement learning algorithms and allows agents to remember and learn from experiences from the past. In an effort to learn more efficiently, researchers proposed prioritized experience replay (PER) which samples important transitions more frequently. In this paper, we propose Prioritized Sequence Experience Replay (PSER) a framework for prioritizing sequences of experience in an attempt to both learn more efficiently and to obtain better performance. We compare the performance of PER and PSER sampling techniques in a tabular Q-learning environment and in DQN on the Atari 2600 benchmark. We prove theoretically that PSER is guaranteed to converge faster than PER and empirically show PSER substantially improves upon PER.
\end{abstract}

\section{Introduction}
Reinforcement learning is a powerful technique to solve sequential decision making problems. Advances in deep learning applied to reinforcement learning resulted in the DQN algorithm \cite{mnih2015human} which uses a neural network to represent the state-action value. With experience replay and a target network, DQN achieved state-of-the-art performance in the Atari 2600 benchmark and other domains at the time.

While the performance of deep reinforcement learning algorithms can be above human-level in certain applications, the amount of effort required to train these models is staggering both in terms of data samples required and wall-clock time needed to perform the training.
This is because reinforcement learning algorithms learn control tasks via trial and error, much like a child learning to ride a bicycle \cite{sutton1998introduction}. In gaming environments, experience is reasonably inexpensive to acquire, but trials of real world control tasks often involve time and resources we wish not to waste. Alternatively, the number of trials might be limited due to wear and tear of the system, making data-efficiency critical \cite{gal2016uncertainty}.
In these cases where simulations are not available or where acquiring samples requires significant effort or expense, it becomes necessary to utilize the acquired data more efficiently for better generalization.

As an important component in deep reinforcement learning algorithms, experience replay has been shown to both provide uncorrelated data to train a neural network and to significantly improve the data efficiency \cite{lin1992self,wang2016sample,zhang2017deeper}. In general, experience replay can reduce the amount of experience required to learn at the expense of more computation and memory \cite{schaul2015prioritized}.

There are various sampling strategies to sample transitions from the experience replay memory. The original primary purpose of the experience replay memory was to decorrelate the input passed into the neural net, and therefore the original sampling strategy was uniform sampling. Prioritized experience replay (PER) \cite{schaul2015prioritized} demonstrated that the agent can learn more effectively from some transitions than from others. By sampling important transitions within the replay memory more often at each training step, PER makes experience replay more efficient and effective than uniform sampling.

In this paper we propose an extension to PER that we term Prioritized Sequence Experience Replay (PSER) that not only assigns high sampling priority to important transitions, but also increases the priorities of previous transitions leading to the important transitions. To motivate our approach, we use the `Blind Cliffwalk' environment introduced in \citet{schaul2015prioritized}. To evaluate our results, we use the DQN algorithm \cite{mnih2015human} with PSER and PER to provide a fair comparison of the sampling strategy on the final performance of the algorithm on the Atari 2600 benchmark. We also prove theoretically that PSER converges faster than PER. Our experimental and theoretical results show using PSER substantially improves upon the performance of PER in both the Blind Cliffwalk environment and the Atari 2600 benchmark.

\section{Related Work}
\subsection{DQN and its extensions}
With the DQN algorithm described in \citet{mnih2015human}, deep learning and reinforcement learning were successfully combined by using a deep neural network to approximate the state-action values, where the input of the neural network is the current state $s$ in the form of pixels, representing the game screen and the output is the state-action values corresponding to different actions (i.e. $Q$-values). It is known that neural networks may be unstable and diverge when applying non-linear approximators in reinforcement learning (RL) algorithms \cite{sutton1998introduction}. DQN uses experience replay and target networks to address the instability issues.
At each time step, based on the current state, the agent selects an action based on some policy (i.e. $\epsilon$-greedily) with respect to the action values, and adds a transition $(s_t, a_t, r_t, s_{t+1})$ to a replay memory. The neural network is then optimized using stochastic gradient descent to minimize the squared TD error of the transitions sampled from the replay memory. The gradient of the loss is back-propagated only into the parameters of the online network and a target network is updated from the online network periodically.

Many extensions to DQN have been proposed to improve its performance.
Double $Q$-learning \cite{van2016deep} was proposed to address the overestimation due to the action selection using the online network.
Prioritized experience replay (PER) \cite{schaul2015prioritized} was proposed to replay important experience transitions more frequently, enabling the agent to learn more efficiently.
Dueling networks \cite{wang2016dueling} is a neural network architecture which can learn state and advantage value, which is shown to stabilize learning.
Using multi-step targets \cite{sutton1998introduction} instead of a single reward is also shown to lead to faster learning.
Distributional RL \cite{bellemare2017distributional} was proposed to learn the distribution of the returns instead of the expected return to more effectively capture the information contained in the value function.
Noisy DQN \cite{fortunato2017noisy} proposed another exploration technique by adding parametric noise to the network weights.

Rainbow \cite{hessel2017rainbow} combined the above mentioned 6 variants together into one agent, achieving better data efficiency and performance on the Atari 2600 benchmark, leading to a new state-of-the-art at the time. Through the ablation procedure described in the paper, the contribution of each component was isolated.
Distributed prioritized replay \cite{horgan2018distributed} utilized a massively parallel approach to show that with enough scaling a new state-of-the-art score can be achieved, but at a cost of orders of magnitude more data. (Typical amounts of frames used for Atari 2600 benchmark games are 200 million frames. The distributed prioritized replay paper has a faster wall-clock time execution, but orders of magnitude more frames were required.)
A comprehensive survey of deep reinforcement learning algorithms including other extensions of DQN can be found at \cite{drl2018survey}.

\subsection{Experience replay}
Experience replay has played an important role in providing uncorrelated data for the online neural network training of deep reinforcement learning algorithms \cite{mnih2015human,lillicrap2015continuous}. There are also studies into how experience replay can influence the performance of deep reinforcement learning algorithms \cite{de2015importance,zhang2017deeper}.

In the experience replay of DQN, the observation sequences are stored in the replay memory and sampled uniformly for the training of the neural network in order to remove the correlations in the data. However, this uniform sampling strategy ignores the importance of each transition and is shown to be inefficient for learning \cite{schaul2015prioritized}.

It is well-known that model-based planning algorithms such as value iteration can be made more efficient by prioritizing updates in an appropriate order. Based on this idea, prioritized sweeping \cite{moore1993prioritized,andre1998generalized} was proposed to update the states with the largest Bellman error, which can speed up the learning for state-based and compact (using function approximator) representations of the model and the value function. Similar to prioritized sweeping, prioritized  experience replay (PER) \cite{schaul2015prioritized} assigns priorities to each transition in the experience replay memory based on the TD error \cite{sutton1998introduction} in model-free deep reinforcement learning algorithms, which is shown to improve the learning efficiency tremendously compared with uniform sampling from the experience replay memory. There are also several other proposed methods trying to improve the sample efficiency of deep reinforcement learning algorithms. \citet{lee2018sample} proposed a sampling technique which updates the transitions backward from a whole episode. \citet{karimpanal2018experience} proposed an approach to select appropriate transition sequences to accelerate the learning.

In another recent study by \citet{Zhong2017RewardBP}, the authors investigate the use of back-propagating a reward stimulus to previous transitions. In our work, we follow the methodology set forth in \citet{schaul2015prioritized} to provide a more general approach by using the current TD error as the priority signal and introduce techniques we found critical to maximize performance.

The approach proposed in this paper is an extension of PER. While we assign a priority to a transition in the replay memory, we also propagate this priority information to previous transitions in an efficient manner, and experimental results on the Atari 2600 benchmark show that our algorithm substantially improves upon PER sampling.

\section{Prioritized Sequence Experience Replay}

\begin{figure}[t]
\begin{center}
\includegraphics[trim={0 0 0 1cm},clip,width=\columnwidth]{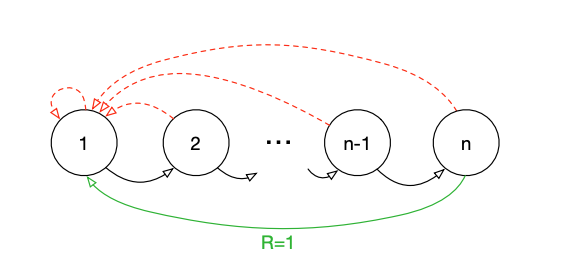}
\caption{The Blind Cliffwalk environment. At each state there are 2 available actions (correct action and wrong action). The agent has to learn to take the correct action at each state to reach the final reward.}

\label{blindcliffwalk_diagram}
\end{center}
\vskip -0.2in
\end{figure}

Using a replay memory is shown to be able to stabilize the neural network \cite{mnih2015human}, but the uniform sampling technique is shown to be inefficient \cite{schaul2015prioritized}. To improve the sampling efficiency, \citet{schaul2015prioritized} proposed prioritized experience replay (PER) to use the last observed TD error to make more effective use of the replay memory for learning. In this paper we propose an extension of PER named Prioritized Sequence Experience Replay (PSER), which can also take advantage of information about the trajectory by propagating the priorities back through the sequence of transitions.

\subsection{A motivating example}
To motivate and understand the potential benefits of PSER, we implemented four different agents in the artificial `Blind Cliffwalk’ environment introduced in \citet{schaul2015prioritized}, shown in Figure~\ref{blindcliffwalk_diagram}. With only $n$ states and two actions, the environment requires an exponential number of random steps until the first non-zero reward; to be precise, the chance that a random sequence of actions will lead to the reward is $2^{-n}$. The first agent replays transitions uniformly from the experience at random, while the second agent invokes an oracle to prioritize transitions, which greedily selects the transition that maximally reduces the global loss in its current state (in hindsight, after the parameter update). The last two agents use PER and PSER sampling techniques respectively. In this chain environment with sparse reward, there is only one non-zero reward that is located at the end of chain marked in green and labelled $R = 1$. The agent must learn the correct sequence of actions in order to reach the goal state and collect the reward. At any state along the chain, the action that leads to the next state varies -- sometimes $a_1$, sometimes $a_2$ -- to prevent the agent from adopting a trivial solution (e.g., always take action $a_1$.)  Any incorrect action results in a terminal state and the agent starts back at the beginning of the chain. For the details of the experiment setup, see the appendix.

\begin{figure*}[t]
\centering
\begin{subfigure}{.45\textwidth}
\centering
\label{subfig:17max}
\includegraphics[trim={0 0 0 2cm},clip,width=\columnwidth]{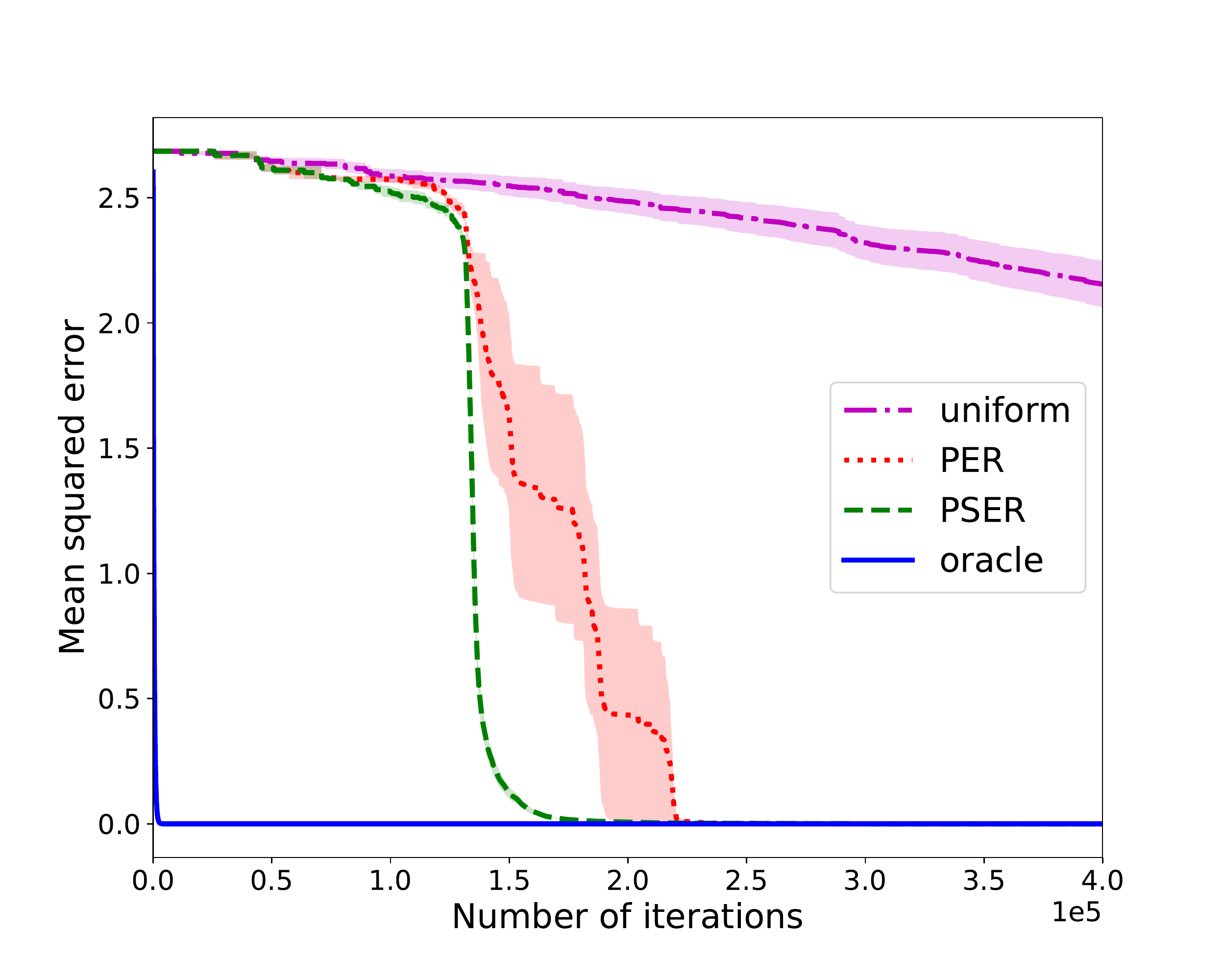}
\caption{16 states with all transitions initialized with max priority.}
\end{subfigure}%
\begin{subfigure}{.45\textwidth}
\centering
\label{subfig:17zero}
\includegraphics[trim={0 0 0 2cm},clip,width=\columnwidth]{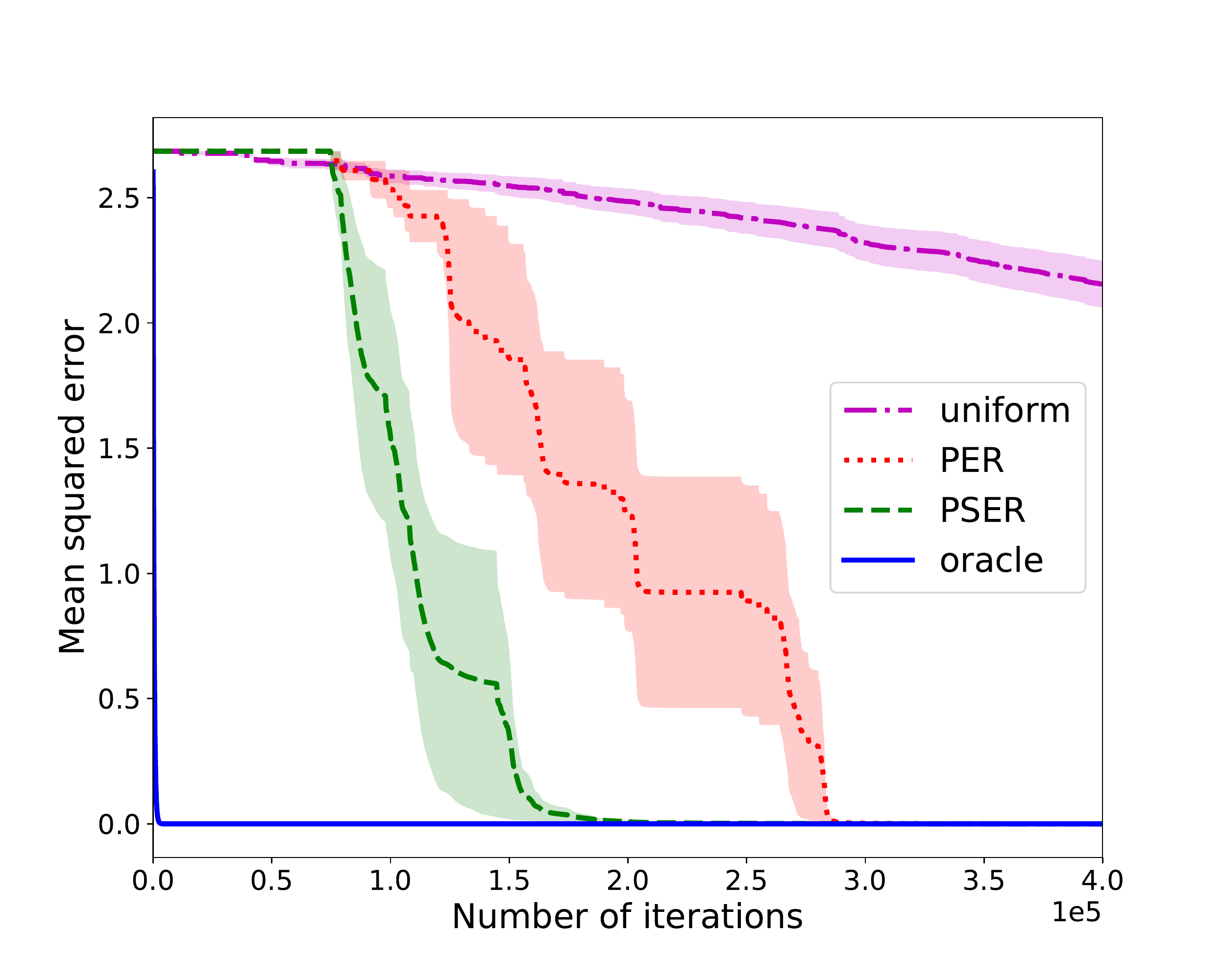}
\caption{16 states with all transitions initialized with $\epsilon$ priority.}
\end{subfigure}
\caption{Comparison of convergence speed for a PSER, PER, uniform, and oracle agent in the Blind Cliffwalk environment with 16 states. PSER shows improved convergence speed as compared to PER and uniform in all cases. The shaded area represents 68\% confidence interval from 10 independent runs with different seeds.}
\label{fig:blindcliff}
\vskip -0.2in
\end{figure*}

To provide some early intuition of the benefits of PSER, we compare performance on the Blind Cliffwalk between PSER and PER. We track the mean squared error between the ground truth $Q$-value and $Q$-learning result every 100 iterations. Better performance in this experiment means that the loss curve more closely matches the oracle. The PER paper demonstrated that by prioritizing transitions based on the TD error, improvements in performance were obtained over uniform sampling. Our results show that by prioritizing the transition with TD error and decaying a portion of this priority to previous transitions, further improvements are obtained with earlier convergence as compared to PER. We show results of this Blind Cliffwalk environment with 16 states in Figure~\ref{fig:blindcliff} and also show how the initialization of the transition's priority (max priority or small non-zero priority, $\epsilon$) in the replay memory affects convergence speed. We find that PSER consistently outperforms PER in this problem.

Examining the curves in Figure~\ref{fig:blindcliff} more closely, there is an initial period where PSER is comparable to both uniform and PER. This is due to all samples initially having the same priority in the replay memory which results in uniform sampling. This uniform sampling continues until the goal transition is sampled from the replay memory and a non-zero TD error is encountered.

At this point, how the agent updates the priority of this transition in the replay memory leads to the divergence in performance between the algorithms. Uniform sampling continues to sample transitions with equal probability from the replay memory. PER updates the priority of the one transition in the memory, but all other transitions in the memory are still chosen at uniform which still results in inefficient sampling as we need to wait until the transition preceding the goal state is sampled. PSER capitalizes on the high TD error that was received and decays a portion of the new priority of the goal state to the preceding states to encourage sampling of the states that led to the goal state. It is clear from Figure~\ref{fig:blindcliff} that by decaying the priority of the high TD error states, we can encourage faster convergence to the true $Q$-value in an intuitive and effective way.

To support this intuition, we offer the following Theorem to describe the convergence rates of PER and PSER (due to space restrictions, the proof can be found in Appendix).

\textbf{Theorem 1} Consider the Blind Cliffwalk environment with $n$ states, if we set the learning rate of the asynchronous $Q$-learning algorithm to 1, then with a pre-filled state transitions in the replay memory by exhaustively executing all $2^n$ possible action sequences, the expected steps for the $Q$-learning algorithm to converge with PER sampling strategy is represented by:
\begin{equation}
    \mathbb{E}_{\text{PER}, n}[N] = 1 + (2^{n+1}-2)(1 - \frac{1}{2^{n-1}})
\end{equation}
and expected steps for the $Q$-learning algorithm to converge with PSER sampling strategy with decaying coefficient $\rho$ is
\begin{equation}
    \mathbb{E}_{\text{PSER}, n}[N] \leq \frac{n}{1-\rho} - \frac{\rho - \rho^{n+1}}{(1 - \rho)^2}
\end{equation}
\vskip -0.1in

In Figure~\ref{th1} we plot the expected number of iterations for convergence from the result of Theorem 1, from which we can see for the Q-learning algorithm, PSER sampling strategy theoretically converges faster.

\subsection{Prioritized sequence decay}
\label{section:sequence_decay}

In this subsection we formally define the concept of the prioritized sequence and decaying priority backward in time within the replay memory.

Formally, the priority of transitions will be decayed as follows:

Suppose in one episode, we have a trajectory of transitions $T_0$ to $T_{n-1}$ ($T_i = (s_i, a_i, r_i, s_{i+1})$) stored in the experience replay memory with priorities $p=(p_0, p_1, \cdots, p_{n-1})$.
If the agent observes a new transition $T_n=(s_n, a_n, r_n, s_{n+1})$, we first calculate its priority $p_n$ based on its TD error similar to the PER algorithm:
\begin{equation}
\label{eq:td_error}
    \delta = r_n + \gamma \max_a Q_{\mathrm{target}} (s_{n+1} , a) - Q(s_n, a_n)
\end{equation}
\vskip -0.20in
\begin{equation}
\label{eq:p_n}
    p_n = | \delta | + \epsilon,
\end{equation}
where $\epsilon$ is a small positive constant to allow transitions with zero TD-error a small probability to be resampled.

As in \citet{schaul2015prioritized}, according to the calculated priority, the probability of sampling transition $i$ is:
\begin{equation}
\label{eq:P_i}
  P(i) = \frac{p_i^\alpha}{\sum_k p_k^\alpha},
\end{equation}

where the exponent $\alpha$ determines how much prioritization is used. We then decay the priority exponentially (with decay coefficient $\rho$) to the previous transitions stored in the replay memory for that episode and apply a $\max$ operator in an effort to preserve any previous priority assigned to the decayed transitions:
\begin{equation}
\label{eq:decay_max}
\begin{split}
p_{n-1} &= \max\{ p_n \cdot \rho, p_{n-1} \}	\\
p_{n-2} &= \max\{ p_n \cdot \rho^2, p_{n-2} \}	\\
p_{n-3} &= \max\{ p_n \cdot \rho^3, p_{n-3} \}	\\
& \quad\quad\quad\cdots
\end{split}
\end{equation}
\vskip -0.1in

\begin{figure*}[t]
\centering
\begin{minipage}{\columnwidth}
  \centering
  \includegraphics[trim={0 0 0 2cm},clip,width=\columnwidth]{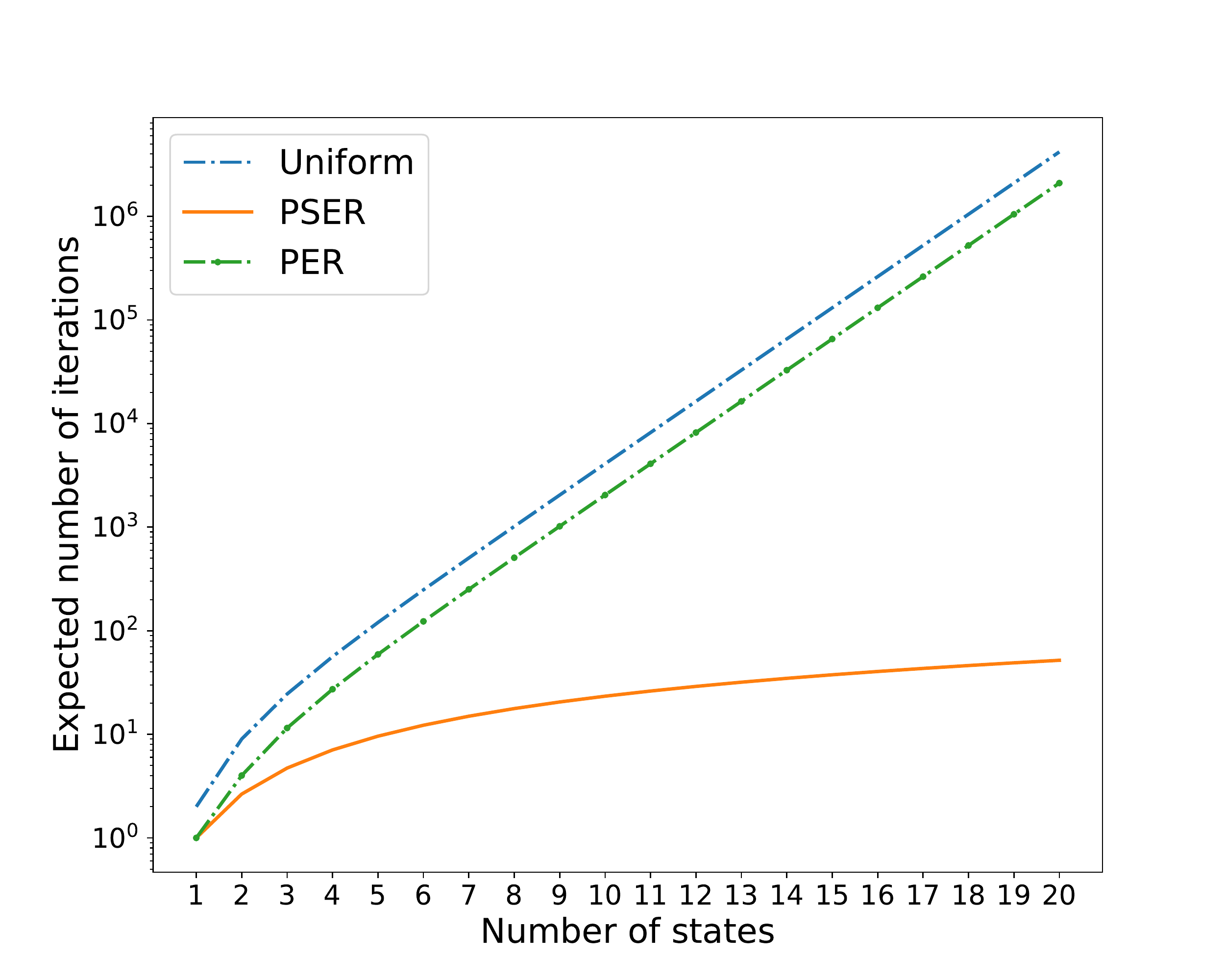}
  \captionof{figure}{From Theorem 1, the expected number of iterations until convergence given the number of states in the Blind Cliffwalk, where lower values along the $y$-axis mean convergence occurs earlier.}
  \label{th1}
\end{minipage}~~~~~~~~%
\begin{minipage}{\columnwidth}
  \centering
  \includegraphics[trim={0 0 0 2cm},clip,width=\columnwidth]{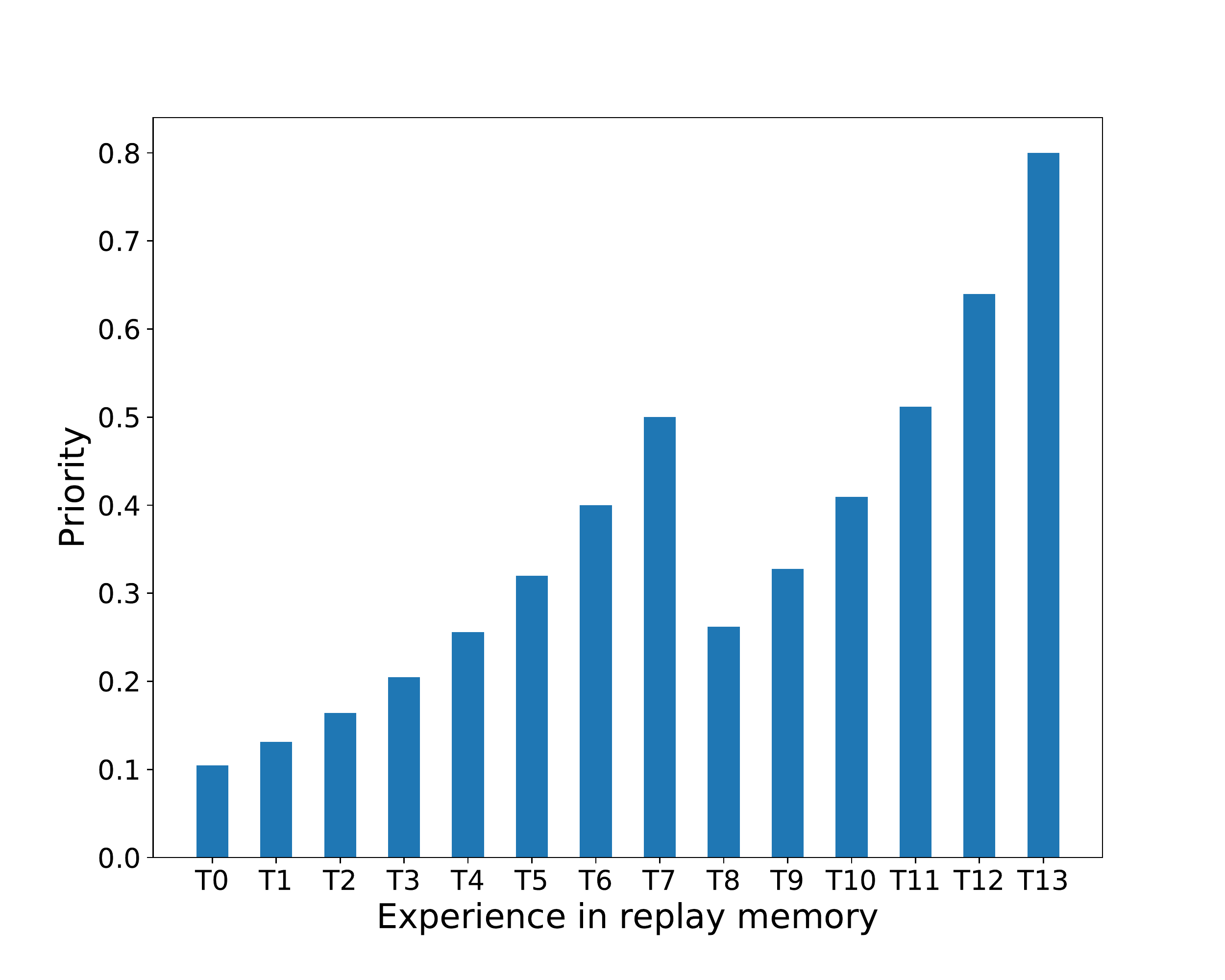}
  \captionof{figure}{A max operator is used to prevent the priority decay due to $T_{13}$ from overwriting a previously calculated priority decay due to $T_7$.\\}
  \label{max_is_good}
\end{minipage}
\vskip -0.2in
\end{figure*}

We refer to this decay strategy as the \textit{MAX} variant. One other potential way to decay the priority is to simply add the decayed priority $p_n \cdot \rho^i$ with the previous priority $p_{n-i}$ assigned to the transition. This we refer to as the \textit{ADD} variant. Note in the \textit{ADD} variant, we keep the priority less than the max priority when decaying the priority backwards, to avoid overflow issues.
\begin{equation}
\label{eq:decay_add}
\begin{split}
p_{n-1} &= \min\{ p_n \cdot \rho + p_{n-1}, \max_n p_n \}	\\
p_{n-2} &= \min\{ p_n \cdot \rho^2 + p_{n-2}, \max_n p_n \}	\\
p_{n-3} &= \min\{ p_n \cdot \rho^3 + p_{n-3}, \max_n p_n \}	\\
& \quad\quad\quad\quad\quad\cdots
\end{split}
\end{equation}
\vskip -0.1 in

Figure~\ref{max_is_good} illustrates this for a case where a priority decay at transition $T_7$ was calculated, then another priority decay occurs at transition $T_{13}$. Without the max operator applied, the priority $p_7$ for transition $T_7$ in the replay memory would be set to $\rho \cdot p_8$ where $p_8$ is the priority for transition $T_8$.

Here we note that as the priority is decayed, we expect that after some number of updates the decayed priority $p_{n-k}$ is negligible and is therefore wasted computation. We therefore define a window of size $W$ over which we will allow the priority $p_n$ to be decayed, after which we will stop. We arbitrarily selected a threshold of $1\%$ of $p_n$ as a cutoff for when the decayed priority becomes negligible. We compute the window size $W$, then, based off the value of the hyperparameter $\rho$ as follows:
\begin{align}
\label{eq:window_size}
    p_n \cdot \rho^W &\leq 0.01 p_{n} \\
    W &\leq \frac{\ln 0.01 }{ \ln \rho}.
\end{align}
Through the above formulation for PSER, we identified an issue which we termed ``priority collapse'' during the decay process.
Suppose for a given environment, PSER has already decayed the priority backward for the ``surprising'' transition, which we will call $T_{i}$. Let's assume that currently all of the $Q$-values are $0$ and we sampled a transition in the replay memory, $T_{i-2}$, that led to $T_{i}$. From Equation~\ref{eq:td_error} and Equation~\ref{eq:p_n} the priority for transition $T_{i-2}$ would drop to $\epsilon$.
The result is that a priority sequence that was recently decayed has almost no effect as it is almost guaranteed to be eliminated at the next sampling. When this happens to multiple states we term this ``priority collapse'' and the potential benefits of PSER are eliminated making it nearly equivalent to traditional PER.

In order to prevent this catastrophic ``priority collapse'' we design a parameter, $\eta$ which forces the priority to decrease slowly. When updating the priority of a sampled transition in the replay memory, we want to maintain a portion of its previous priority to prevent it from decreasing too quickly:
\begin{equation}
  p_i \leftarrow \max (| \delta |+\epsilon, \eta \cdot p_i).
\end{equation}
where $i$ here refers to the index of the sampled transition within the replay memory. Without this decay parameter $\eta$, we experimentally found PSER to have no significant benefit over PER which confirms our intuition.

Our belief is that the decay parameter $\eta$ provides time for the Bellman update process to propagate information about the TD error through the sequence and for the neural network to more readily learn an appropriate $Q$-value approximation.

\subsection{Annealing the bias}

As discussed in \citet{schaul2015prioritized}, prioritized replay introduces bias because it changes the sampling distribution of the replay memory. To correct the bias, \citet{schaul2015prioritized} introduced importance-sampling (I.S.) weights defined as follows:
\begin{equation}
    w_{i} = \bigg(\frac{1}{N}\cdot\frac{1}{P(i)}\bigg)^{\beta},
\end{equation}
where $N$ is the size of the replay memory and $P(i)$ is the probability of sampling transition $i$. Non-uniform probabilities are fully compensated for if $\beta = 1$. In PSER, we adapt the I.S. weights to correct for the sampling bias. The full algorithm is presented in Algorithm \ref{alg:algorithm1} in the Appendix.

\section{Experimental Methods}

\begin{figure*}[t]
\begin{center}
\includegraphics[width=2\columnwidth,height=9cm]{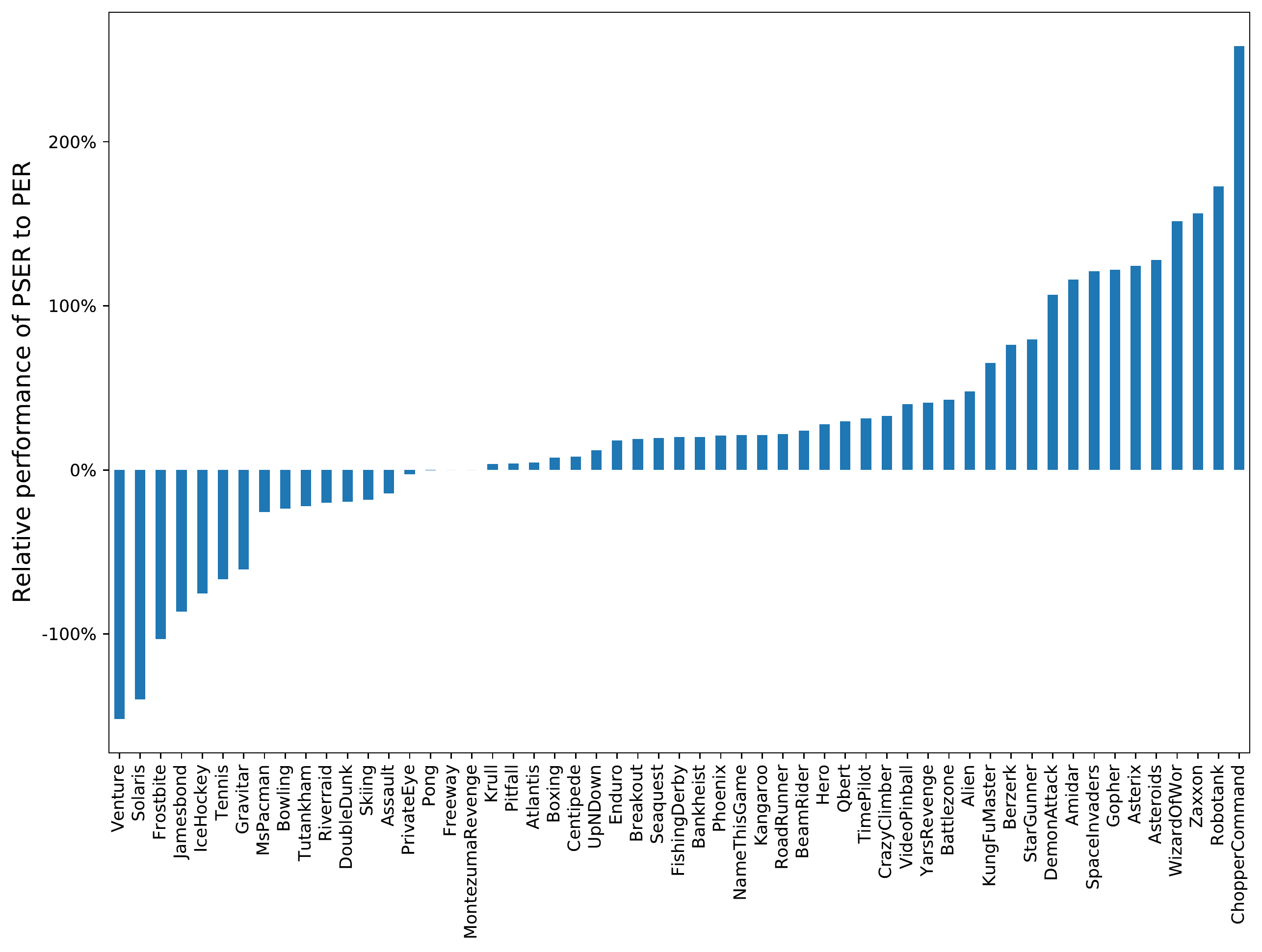}
\caption{Relative performance of prioritized sequence experience replay (PSER) to prioritized experience replay (PER) in all 55 Atari 2600 benchmark games where human scores are available. 0\% on the vertical axis implies equivalent performance; positive numbers represent the cases where PSER performed better; negative numbers represent the cases where PSER performed worse.}
\label{bar_plot}
\end{center}
\vskip -0.2in
\end{figure*}

\subsection{Evaluation methodology}
We used the Arcade Learning Environment \cite{bellemare2013arcade} to evaluate the performance of our proposed algorithm. We follow the same training and evaluation procedures of \citet{hessel2017rainbow,mnih2015human,van2016deep}. We calculate the average score during training every 1M frames in the environment. After every 1M frames, we then stop training and evaluate the agent's performance for 500K frames. We also truncate the episode lengths to 108K frames (or 30 minutes of simulated play) as in \citet{van2016deep, hessel2017rainbow}.
In the results section, we report the mean and median human normalized scores of PSER and PER in the Atari 2600 benchmark and in the appendix we provide full learning curves for all games in the \textit{no-op starts} testing regime.

\subsection{Hyperparameter tuning}

DQN has a number of different hyperparameters that can be tuned.
To provide a comparison with our baseline DQN agent, we used the hyperparameters that are provided in \citet{mnih2015human} for the DQN agent formulation (see Appendix for more details).

Our PSER implementation also has hyperparameters that require tuning. Due to the large amount of time it takes to run the full 200M frames for the DQN tests (multiple days), we used a coordinate descent approach to tune the PSER parameters for a subset of the Atari 2600 benchmark.
In the coordinate descent approach, we define a set of different values to test for each parameter. Then, holding all other parameters constant, we tune one parameter until the best result is obtained. We then fix this tuned parameter and move to the next parameter and repeat this process until all parameters have been tuned. While this does not test every combination of parameter values, it greatly reduces the hyperparameter search space and proved to provide good results.

The hyperparameters obtained during the hyperparameter search were used for all Atari 2600 benchmark results reported in this paper. Hyperparameters were not tuned for each game so as to better measure how the algorithm generalizes over the whole suite of the Atari 2600 benchmark. We found the best results were obtained with $W$ = 5, $\rho$ = 0.4, and $\eta$ = 0.7.
\begin{figure*}[t]
\begin{center}
\includegraphics[trim={0 0 0 2cm},clip,width=2\columnwidth,height=9cm]{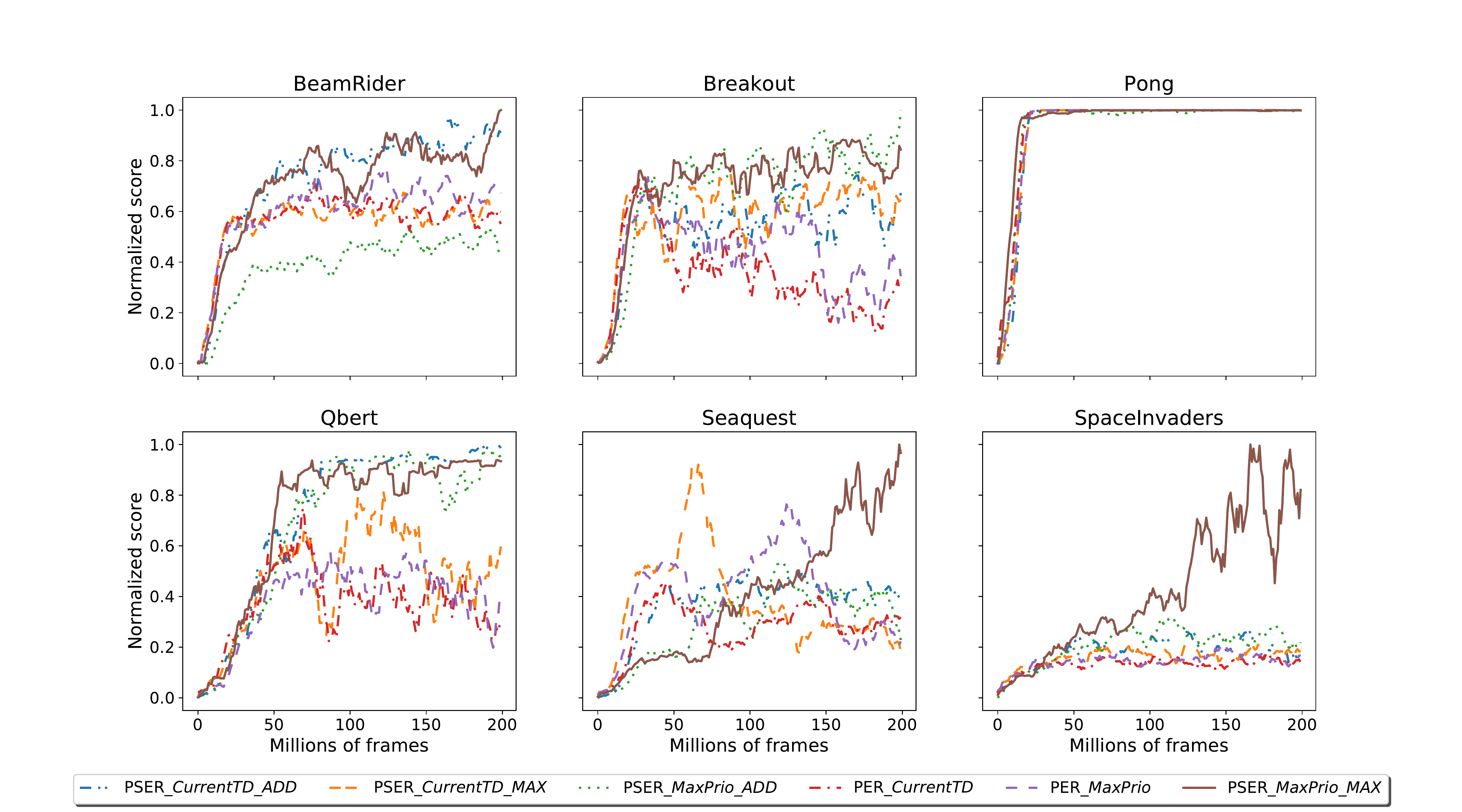}
\caption{Ablation study performed on six Atari 2600 games. We show the full learning curves from the evaluation period that occurs following each 1M frames of training. Scores are normalized by the maximum and minimum value recorded across all ablations for each game. The legend is read as Sampling Strategy\_Initial Prioritization\_Decay Scheme. For example, PSER\_\textit{CurrentTD}\_\textit{MAX} corresponds to the learning curve for PSER, \textit{CurrenTD} initial prioritization, and \textit{MAX} decay strategy. Results are smoothed with a 10M frame rolling average to improve clarity.}
\label{ablation_study}
\end{center}
\vskip -0.2in
\end{figure*}
\section{Analysis}
In this section, we analyze the main experimental results using the Atari 2600 benchmark available within the OpenAI gym environment \cite{brockman2016openai}. We show that by adding PSER to the DQN agent we can achieve substantial improvement to performance as compared to PER.

\subsection{Baselines}
We compared PSER to PER using the version of DQN described in \citet{mnih2015human}. This way we can provide a fair comparison by minimally modifying the algorithm to attribute any performance differences to the sampling strategy. Both DQN agents used identical hyperparameters that we list in the Appendix.

\subsection{Comparison with baselines}

Figure~\ref{bar_plot} shows the relative performance of prioritized sequence experience replay and prioritized replay for the 55 Atari 2600 games for which human scores are available.
We compute the human normalized scores for PER and PSER following the methodology in \citet{schaul2015prioritized} which we repeat in the Appendix for clarity. A comparison of all 60 games showing percent improvement of PSER over PER is also available in the Appendix.

We can see from Figure~\ref{bar_plot} that PSER leads to substantial improvements over PER. In the games where PSER outperformed PER, we can see that the range of relative difference is much larger as compared to the games where PER outperformed PSER. For PSER, 8 games achieved a relative difference of over 100\% as compared to 3 for PER.

In Table~\ref{table_atari} we compare the final evaluation performance of PSER and PER on the Atari 2600 benchmark by calculating the median and mean human normalized scores (See the appendix for the learning curves of all Atari games). PSER achieves a median score of $109\%$ and a mean score of $832\%$ in the no-ops regime, significantly improving upon PER.
\begin{table}[H]
\caption{Median and Mean human normalized scores of the best agent snapshot across 55 Atari games for which human scores are available.}
\label{table_atari}
\begin{center}
\begin{small}
\begin{sc}
\begin{tabular}{lcccr}
\toprule
Sampling Strategy & Median & Mean \\
\midrule
PSER        & 109\%   &  832\%      \\
PER         & 88\%  &  607\%    \\
\bottomrule
\end{tabular}
\end{sc}
\end{small}
\end{center}
\end{table}
\subsection{Ablation Study}
To understand how the initial priority assigned to a transition interacts with prioritized sampling, we conducted additional experiments to evaluate the performance.

There are two variants of initial priority assignment that we considered in our ablation study. First, in \citet{mnih2015human, van2016deep, hessel2017rainbow}, transitions are added to the replay memory with the maximum priority ever seen. Second, in \citet{horgan2018distributed} transitions are added with priority calculated from the current TD error of the online model. We refer to the variants as \textit{MaxPrio} and \textit{CurrentTD}, respectively.

In each ablation study, we test combinations of the following parameters: a) prioritized sampling strategy (PSER, PER), b) the initial priority assignment (\textit{MaxPrio}, \textit{CurrentTD}), and c) decay scheme\footnote{The decay scheme is unique to PSER, so this is not tested for PER.} (\textit{MAX}, \textit{ADD}) as described in (6) and (7).

Figure~\ref{ablation_study} compares the performance across six Atari 2600 games. We can see that the choice of the initial priority assignment doesn't appear to lead to a substantial difference in the initial learning speed or performance for both PSER and PER in each game except Seaquest. In Seaquest, we observe that the \textit{CurrentTD} variant leads to faster learning in the initial 75M frames, but begins to hurt performance throughout the remainder of training, potentially due to over-fitting.

We also find that the \textit{MAX} decay strategy led to better performance over the \textit{ADD} decay strategy. Intuitively, to help encourage the Bellman update process from states with high TD error, it makes sense to decay the priority exponentially backwards instead of adding the priorities together.

\subsection{Learning Speed}
Each agent is run on a single GPU and the learning speed for each variant varies depending on the game. For a full 200 million frames of training, this corresponded to approximately 5-10 days of computation time depending on the hardware used\footnote{We adapted the Dopamine \citep{dopamine} code-base for PSER and PER to compare performance.}. We found that the learning speed of PSER is comparable to PER when a small decay coefficient value is used. As this value increases, there is an increase in the computation time due to the larger decay window.

\section{Discussion}
We have demonstrated that PSER achieves substantial performance increases through the Blind Cliffwalk environment and the Atari 2600 benchmark.

While performing this analysis, we tested different configurations of PSER and discovered phenomena that we did not expect. Most important was the priority collapse issue described in section \ref{section:sequence_decay}. By introducing the parameter $\eta$ to maintain a portion of a transition's previous priority, we prevent the altered priorities created by PSER from quickly reverting to the priorities assigned by PER. We believe that there are two processes inherent in deep reinforcement learning: the Bellman update process inherent in all reinforcement learning and Markov Decision Processes, and the neural network gradient descent update process. Both processes are very slow and require many samples to converge.  We hypothesize that keeping the previous transitions' priorities elevated in the replay memory results in additional Bellman updates for the sequence with valuable information. While this speeds up the Bellman update process, it also serves to provide the neural net with better targets which improves the overall convergence rate.

When running experiments on the Atari 2600 benchmark, we needed to choose a fixed hyperparameter set for a fair comparison between PSER and PER. However, the games in the Atari 2600 benchmark vary in how to obtain reward and how long the delay is between action and reward. Even though we achieved substantial improvement over PER with a fixed decay window, allowing the decay window to vary for each game may lead to better performance. One approach is to introduce an adaptive decay window based off the magnitude of the TD error. We leave the investigation of an adaptive decay window to future work.

It remains unclear whether the \textit{MaxPrio} or \textit{CurrentTD} initial priority assignment should be used when adding new transitions to the replay memory. For the Blind Cliff Walk experiments, we found that \textit{MaxPrio} approach delayed the convergence to the true $Q$-value as compared to \textit{CurrentTD}. However on Atari, we found the \textit{MaxPrio} approach to be more effective. Intuitively, adding transitions to the replay memory with the current TD error makes sense to encourage the agent to initially sample these high priority transitions sooner. Adding with max priority should result in an artificially high priority for most new transitions.
We hypothesize that this may be related to the priority collapse problem where these artificially high priorities are temporarily allowing better information flow during the learning process.

We chose to implement PSER on top of DQN primarily for the purpose of enabling a fair comparison in the experiments between PER and PSER, but combining PSER with other algorithms is an interesting direction for future work. For example, PSER can also be used with other off-policy algorithms such as Double $Q$-learning and Rainbow.

\section{Conclusion}
In this paper we introduced Prioritized Sequence Experience Replay (PSER), a novel framework for prioritizing sequences of transitions to both learn more efficiently and effectively. This method shows substantial performance improvements over PER and Uniform sampling in the Blind Cliffwalk environment, and we show theoretically that PSER is guaranteed to converge faster than PER. We also demonstrate the performance benefits of PSER in the Atari 2600 benchmark with PSER outperforming PER in 40 out of 60 Atari games. We show that improved ability for information to flow during the training process can lead to faster convergence, as well as, increased performance, potentially leading to increased data efficiency for deep reinforcement learning problems.

\newpage

\section*{Appendix}
\setcounter{section}{0}
\section{Theorem}
We define the Blind Cliffwalk as the following Markov Decision Process (MDP). The state space of this MDP is composed by $n$ different state: $\{ s_1, s_2, \cdots, s_n \}$. At each state, the agent has two actions to choose $\{a_1, a_2\}$ and here we assume $a_1$ is the correct action and $a_2$ is the wrong action. Correct action will take the agent to next state and wrong action will take the agent back to the initial state $s_1$:
\begin{equation}
    T(s_i, a_j) =
    \begin{cases}
    s_{i+1}, & \mbox{for $i \in \{ 1, \cdots, n-1 \}, j=1$,}  \\
    s_1, & \mbox{otherwise}
  \end{cases}
\end{equation}
The reward function is defined where the agent can get positive reward $r$ only from taking the correct action from state $s_n$:
\begin{equation}
    R(s_i, a_j) =
    \begin{cases}
    1, & \mbox{if $i=n$ and $j=1$}  \\
    0, & \mbox{otherwise}
    \end{cases}
\end{equation}

The $Q$-learning algorithm \cite{watkins1992q} estimates the state-action value function (for discounted
return) as follows:
\begin{equation}
\label{Qlearning}
\begin{split}
    Q_{t+1}(s, a)=&(1-\alpha_{t}) Q_{t}(s, a)\\
&+\alpha_{t}(R(s, a)+\gamma \max _{b \in U(s')} Q_{t}(s', b))
\end{split}
\end{equation}
where $s'$ is the state reached from state $s$ when performing action $a$ at time $t$, and $\alpha$ is the learning rate of the $Q$-learning algorithm at time $t$.

In this paper we consider an asynchronous $Q$-learning process which updates a single entry at each step with different sampling strategies (PER and PSER) from the experience replay memory.

Since the Blind Cliffwalk environment is a deterministic world, we can set the learning rate of the $Q$-learning algorithm to 1, which means after one update we can get the accurate $Q$-value. Next we present the convergence speed of $Q$-learning algorithm with PER and PSER sampling strategy, showing PSER sampling strategy can help $Q$-learning algorithm converge much faster than PER sampling strategy.

\textbf{Theorem 1} Consider the Blind Cliffwalk environment with $n$ states, if we set the learning rate of the asynchronous $Q$-learning algorithm in Equation~\ref{Qlearning} to 1, then with a pre-filled state transitions in the replay memory by exhaustively executing all $2^n$ possible action sequences, the expected steps for the $Q$-learning algorithm to converge with PER sampling strategy is
\begin{equation}
    \mathbb{E}_{\text{PER}, n}[N] = 1 + (2^{n+1}-2)(1 - \frac{1}{2^{n-1}})
\end{equation}
and expected steps for the $Q$-learning algorithm to converge with PSER sampling strategy with decaying coefficient $\rho$ is
\begin{equation}
    \mathbb{E}_{\text{PSER}, n}[N] \leq \begin{cases}
    \frac{n(n+1)}{2}, & \mbox{if $\rho = 0.5$}  \\
    \frac{n}{1-2\rho} - \frac{2\rho - (2\rho)^{n+1}}{(1 - 2\rho)^2}, & \mbox{otherwise}
    \end{cases}
\end{equation}

\textbf{Proof} We first define the ``$Q$-interval'' for the $Q$-learning process, then we show after $n$ $Q$-intervals, the $Q$-learning algorithm will be guaranteed to converge to the true $Q$-value, finally we calculate the expected steps of each $Q$-interval for PER and PSER sampling strategies.

Here we define a ``$Q$-interval'' to be an interval in which every state-action pair $(s, a)$ is tried at least once. Without loss of generality, we initialize the $Q$-value for each state-action pair to be 0 (e.g., $\hat{Q}_0(s,a) = 0, \forall{s, a}$), and we denote the $Q$-value of the $Q$-learning algorithm after $i$th $Q$-interval as $\hat{Q}_i(s,a)$.

Then we show the $Q$-learning algorithm will converge to the true $Q$-value after $n$ $Q$-intervals by induction. From value iteration we know the true $Q$-value function has the following form:
\begin{equation}
    Q^*(s_i, a_j) =
    \begin{cases}
    \gamma^{n-i}, & \mbox{for $j=1$}  \\
    0, & \mbox{otherwise}
    \end{cases}
\end{equation}
The base case is after the first $Q$-interval, we have \begin{equation}
    \hat{Q}_i(s, a) = Q^*(s, a), \forall s \in \{ s_n \}, a \in \{ a_1, a_2 \}
\end{equation}
This is true since we have
\begin{equation}
\begin{split}
    \hat{Q}_1(s_n, a_1) =& R(s_n, a_1) + \gamma \max_{a'} \hat{Q}_0(s_1, a')   \\
    =& 1    \\
    =& \gamma^{n-n}
\end{split}
\end{equation}
and
\begin{equation}
   \hat{Q}_1(s_n, a_2) = R(s_n, a_1) + \gamma \max_{a'} \hat{Q}_0(s_1, a') = 0
\end{equation}

Assume after $i$th $Q$-interval, we have
\begin{equation}
    \hat{Q}_i(s, a) = Q^*(s, a), \forall s \in \{ s_{n-i+1}, \cdots, s_n \}, a \in \{ a_1, a_2 \}
\end{equation}
Then after the $(i+1)$th $Q$-interval, we have
\begin{equation}
\label{1induc1}
\begin{split}
    \hat{Q}_{i+1}(s_{n-i}, a_1) =& R(s_{n-i}, a_1) + \gamma \max_{a'} \hat{Q}_i(s_{n-i+1}, a')   \\
    =& 0 + \gamma \times \max_{a'} Q^*(s_{n-i+1}, a')    \\
    =& \gamma \times \max \{ \gamma^{i-1}, 0 \} \\
    =& \gamma^{i} \\
    =& Q^*(s_{n-i})
\end{split}
\end{equation}
and
\begin{equation}
\label{1induc2}
   \hat{Q}_{i+1}(s_{n-i}, a_2) = R(s_{n-i}, a_2) + \gamma \max_{a'} \hat{Q}_0(s_1, a') = 0
\end{equation}
Also, for $s \in \{ s_{n-i+1}, \cdots, s_n \}$, their values $\hat{Q}_{i+1}(s, a)$ will not change since
\begin{equation}
\label{1induc3}
\begin{split}
    \hat{Q}_{i+1}(s, a) =& R(s, a) + \gamma \max_{a'} \hat{Q}_i(s', a') \\
    =& R(s, a) + \gamma \max_{a'} Q^*(s', a') \\
    =& Q^*(s, a)
\end{split}
\end{equation}
where we use the Bellman optimality equation.

Thus from Equations~\ref{1induc1},\ref{1induc2},\ref{1induc3} we conclude that after the $n$th $Q$-interval,
\begin{equation}
    \hat{Q}_i(s, a) = Q^*(s, a), \forall s \in \{ s_1, \cdots, s_n \}, a \in \{ a_1, a_2 \}
\end{equation}

Finally we calculate the expected steps of each interval for PER and PSER sampling strategy. For a fair comparison between PER sampling strategy and PSER sampling strategy, the replay memory for sampling is first filled with state transitions by exhaustively executing all $2^n$ possible sequences of actions until termination (in random order), in this way the total number of state transitions will be $2^{n+1} - 2$. While initializing the replay memory, each transition will be assigned a priority equal to the TD error of the transition. After the initialization, only the state transition $(s_n, a_1, r, s_1)$ has priority 1 and all the remaining states transitions have priority 0 (here we keep the priorities to be 0 instead of a small number $\epsilon$ for simplicity). When performing the $Q$-learning iteration, the state transitions are sampled according to their priorities.

In fact, from the induction process we can see the expected number of steps in the $i$th $Q$-interval (denoted as $N_i$) equals the expected number for state-action $(s_{n-i+1}, a_1)$ to be sampled. We use this insight to calculate $\mathbb{E}[N_i]$.

For PER sampling strategy,
\begin{equation}
\begin{split}
    \mathbb{E}[N_1] &= 1 \\
    \mathbb{E}[N_i] &= \frac{2^{n+1} - 2}{2^{i-1}}, \text{ for } i \in \{ 2, \cdots, n \}
\end{split}
\end{equation}
The first equation is immediately from the fact that only transition $(s_n, a_1, r, s_1)$ has non-zero priority (whose priority after updating drops to 0). The second equation follows from the fact that there are $2^{i-1}$ state transitions $(s_{n-i+1}, a_1, r, s_{n-i})$. After the $(i-1)$th $Q$-interval, all transitions have equal priority and the probability for transition $(s_{n-i+1}, a_1, r, s_{n-i})$ to get sampled equals
\begin{equation}
    p = \frac{2^{i-1}}{2^{n+1} - 2}
\end{equation}
Thus the expected number of steps for PER sampling strategy to converge equals
\begin{equation}
\label{per_iter_steps}
\begin{split}
    \mathbb{E}_{\text{PER}, n}[N] &= \sum_{i=1}^n \mathbb{E}[N_i]\\
    &= 1 + \frac{2^{n+1}-2}{2} + \cdots + \frac{2^{n+1}-2}{2^{n-1}} \\
    &= 1 + (2^{n+1}-2)(1 - \frac{1}{2^{n-1}})
\end{split}
\end{equation}
We can see that as $n\rightarrow \infty$, $\mathbb{E}_{\text{PER}, n}[N] \rightarrow 2^{n+1}$, which indicates the number of steps to convergence will grow exponentially.

Next we consider PSER sampling strategy. When initializing the replay memory, the PSER sampling strategy will assign the state transition $(s_n, a_1, r, s_1)$ priority 1 according to its TD error. Then PSER decays the priority backwards according to decay coefficient $\rho \in (0,1)$ so that transition $(s_{n-1}, a_1, r, s_{n})$ has priority $\rho$, transition $(s_{n-i+1}, a_1, r, s_{n-i})$ has priority $\rho^{i-1}$. Thus in the first $Q$-interval, let $N_1$ denote the expected number of steps to sample transition $(s_n, a_1, r, s_1)$, let $A_k$ denote the event that transition $(s_{n-1}, a_1, r, s_{n})$ gets sampled at the $k$th sample where $k \in (0, \infty)$, then we have
\begin{equation}
\begin{split}
    \mathbb{E}[N_1] &= \sum_{k=0}^\infty P(A_k) k   \\
    &= P(A_1) \times 1 + P(A_2) \times 2 + \cdots   \\
    &\leq p \times 1 + (1-p)p \times 2 + \cdots \\
    &= 1/p  \\
    &= \sum_{i=1}^n \rho^{n-i}
\end{split}
\end{equation}
where $p = P(A_1) = \frac{1}{\sum_{i=1}^n (\rho^{n-i})}$ and the inequality is due to the fact that if we samples other state transitions, their priority will drop and the probability of transition $(s_{n-1}, a_1, r, s_{n})$ gets sampled will increase. Similarly, we have
\begin{equation}
\begin{split}
    \mathbb{E}[N_i] &\leq \frac{ \sum_{k=i-1}^{n-1} \rho^k }{\rho^{i-1}}    \\
    &= \sum_{k=0}^{n-i} \rho^k
\end{split}
\end{equation}
Thus, we have
\begin{equation}
\label{pser_iter_steps}
\begin{split}
    \mathbb{E}_{\text{PSER}, n}[N] &= \sum_{i=1}^n \mathbb{E}[N_i]\\
    &\leq \sum_{i=1}^n \sum_{k=0}^{n-i} \rho^{k} \\
    &= \frac{n}{1-\rho} - \frac{\rho - \rho^{n+1}}{(1 - \rho)^2}
\end{split}
\end{equation}

We can see that as $n\rightarrow \infty$, $\mathbb{E}_{\text{PSER}, n}[N] \rightarrow n$,
which indicates the number of steps to convergence will grow linearly with $n$. Therefore the PSER sampling strategy converges much faster than the PER in Blind Cliffwalk. $\blacksquare$

Next we show for PSER sampling strategy, the expected steps in one Q-interval are fewer than PER sampling strategy. In fact, calculating the expected steps for each Q-interval is intractable since the priorities keep changing throughout the sampling process. So here we consider the expected steps in one Q-interval for any given priorities.

\section{Blind Cliffwalk Experiment}
For the Blind Cliffwalk experiments, we use a tabular $Q$-learning setup with four different experience replay scheme, where the $Q$-values are represented using a tabular look-up table.

For the tabular $Q$-learning algorithm, the replay memory of the agent is first filled by exhaustively executing all $2^n$ possible sequences of actions until termination (in random order). This guarantees that exactly one sequence will succeed and hit the final reward, and all others will fail with zero reward. The replay memory contains all the relevant experience (the total number of transitions is $2^{n+1} - 2$) at the frequency that it would be encountered when acting online with a random behavior policy.

After generating all the transitions in the replay memory, the agent will next select a transition from the replay memory to learn at each time step. For each transition, the agent first computes its TD-error using:
\begin{equation}
\delta_{t} :=R_{t}+\gamma_{t} \max _{a} Q(S_{t}, a)-Q(S_{t-1}, A_{t-1})
\end{equation}
and updates the parameters using stochastic gradient ascent:
\begin{equation}
\theta \leftarrow \theta+\eta \cdot \delta_{t} \cdot \nabla_{\theta}\left.Q\right|_{S_{t-1}, A_{t-1}}=\theta+\eta \cdot \delta_{t} \cdot \phi(S_{t-1}, A_{t-1})
\end{equation}

The four different replaying schemes we will be using here are uniform, oracle, PER, and PSER. For uniform replaying scheme, the agent will randomly select the transition from the replay memory uniformly. For the oracle replaying scheme, the agent will greedily select the transition that maximally reduces the global loss (in hindsight, after the parameter update). For the PER replaying scheme, the agent will first set the priorities of all transitions to either 0 or 1. Then after each update, the agent will assign new priority to the sampled transition using:
\begin{equation}
    p = |\delta| + \epsilon
\end{equation}
and the probability of sampling transition $i$ is
\begin{equation}
    P(i) = \frac{p_i^\alpha}{\sum_k p_k^\alpha}
\end{equation}
where $\delta$ is the TD error for the sampled transition which can be calculated from equation (1), $\alpha = 0.5$ and $\epsilon = 0.0001$. For the PSER replaying scheme, we first calculate the priority as in equation (3) and propagate back the priority 5 steps before:
\begin{equation}
    \label{eq:decay}
    \begin{split}
    p_{n-1} &= \max\{\rho^1 p_n, p_{n-1}\}	\\
    p_{n-2} &= \max\{\rho^2 p_n, p_{n-2}\}	\\
    p_{n-3} &= \max\{\rho^3 p_n, p_{n-3}\}  \\
    & \cdots
    \end{split}
\end{equation}
Then the agent will sample transitions from the replay memory with probability based on Equation~\ref{eq:P_i}.

For this experiment, we vary the size of the problem (number of states $n$) from 13 to 16. The discount factor is set to $\gamma = 1 - \frac{1}{n}$ which keeps values on approximately the same scale independently of $n$. This allows us to use a fixed step-size of $\eta = \frac{1}{4}$ in all experiments.

\begin{figure*}[p]

\centering

\subcaptionbox{
13 states with all transitions initialized with max priority.
\label{subfig:13maxappen}
}{
\includegraphics[width=0.7\columnwidth]{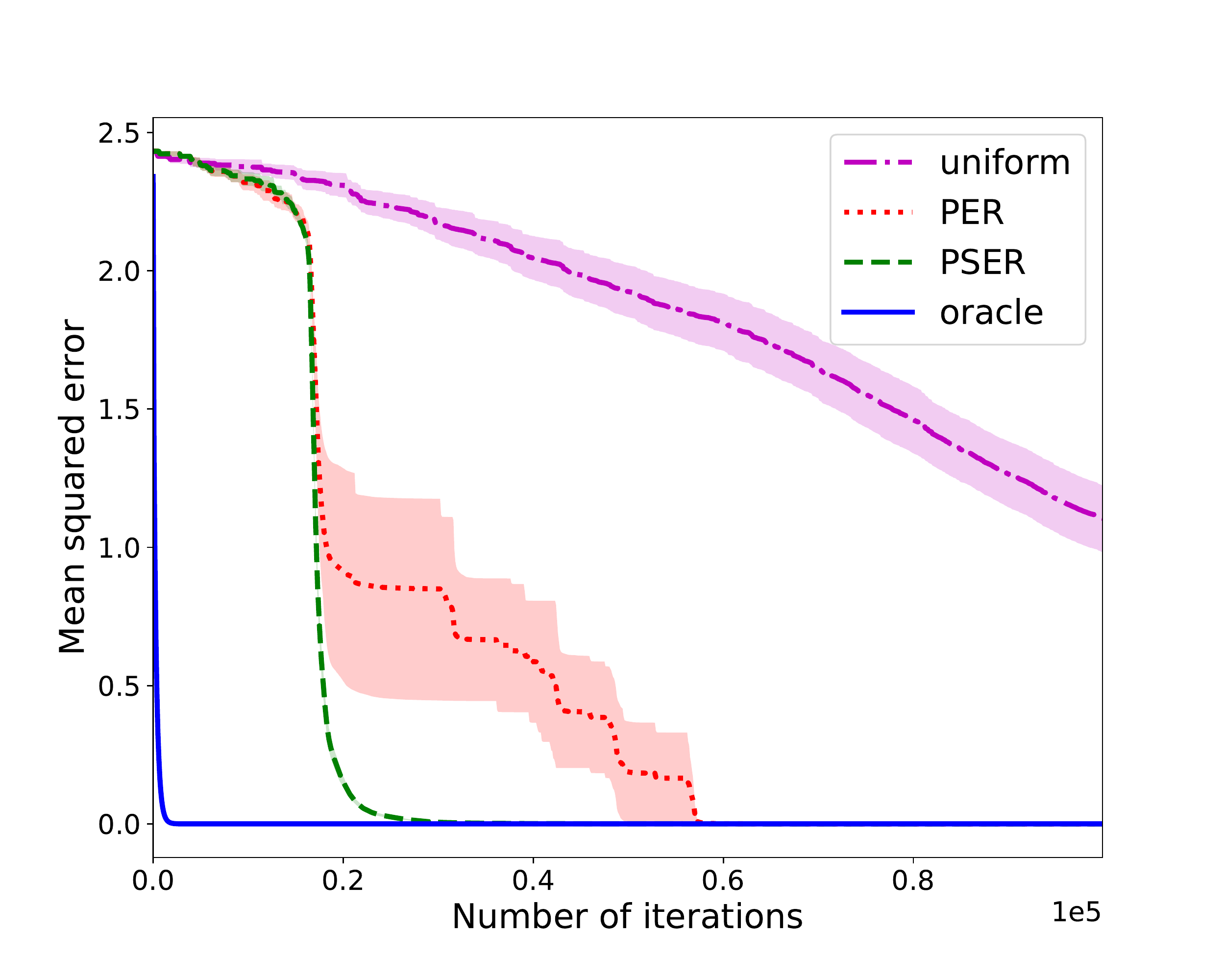}
}%
\hspace{0.08\textwidth} 
\subcaptionbox{
13 states with all transitions initialized with $\epsilon$ priority.
\label{subfig:13zeroappen}
}{
\includegraphics[width=0.7\columnwidth]{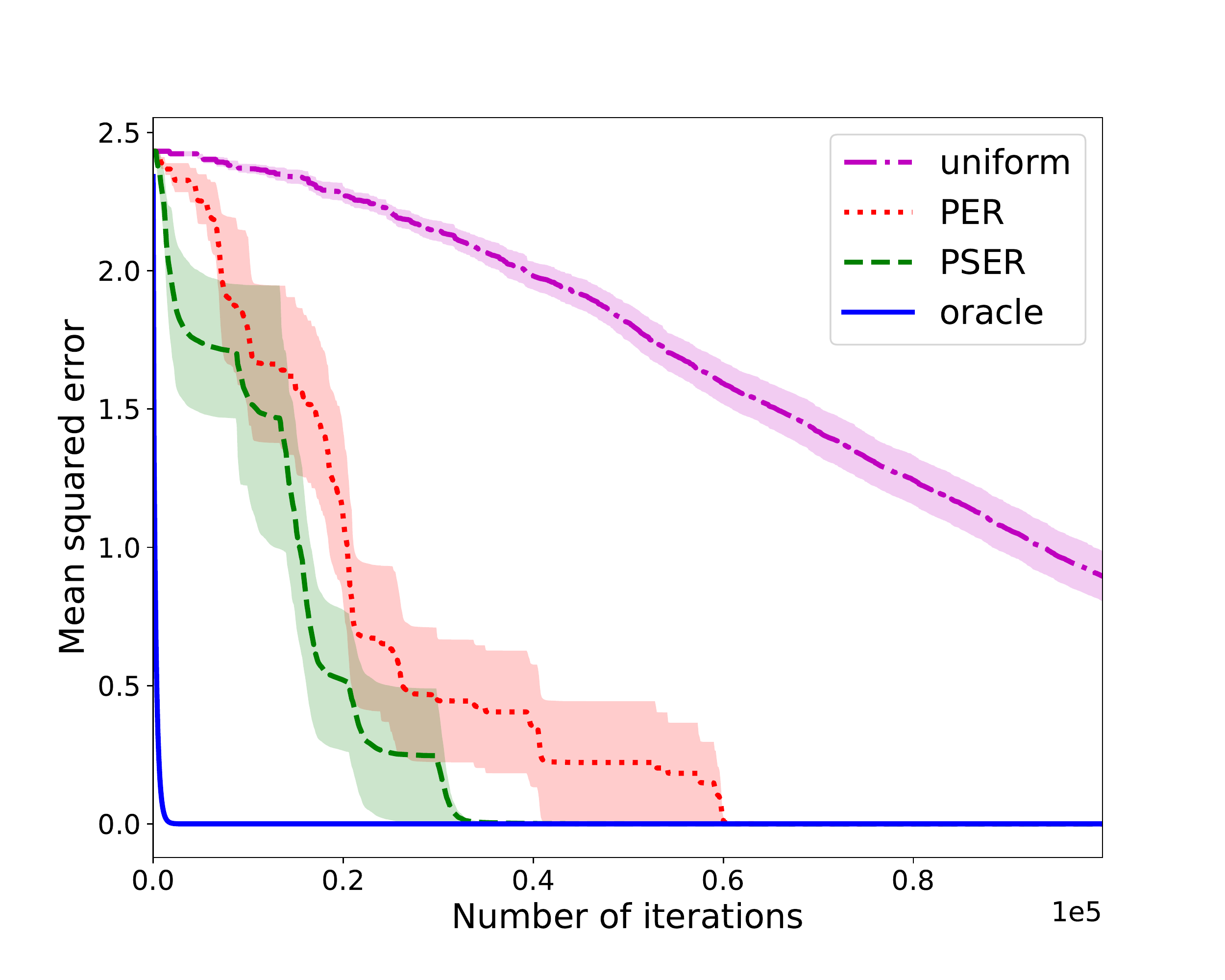}
}%
\hspace{0.08\textwidth} 
\subcaptionbox{
14 states with all transitions initialized with max priority.
\label{subfig:14maxappen}
}{
\includegraphics[width=0.7\columnwidth]{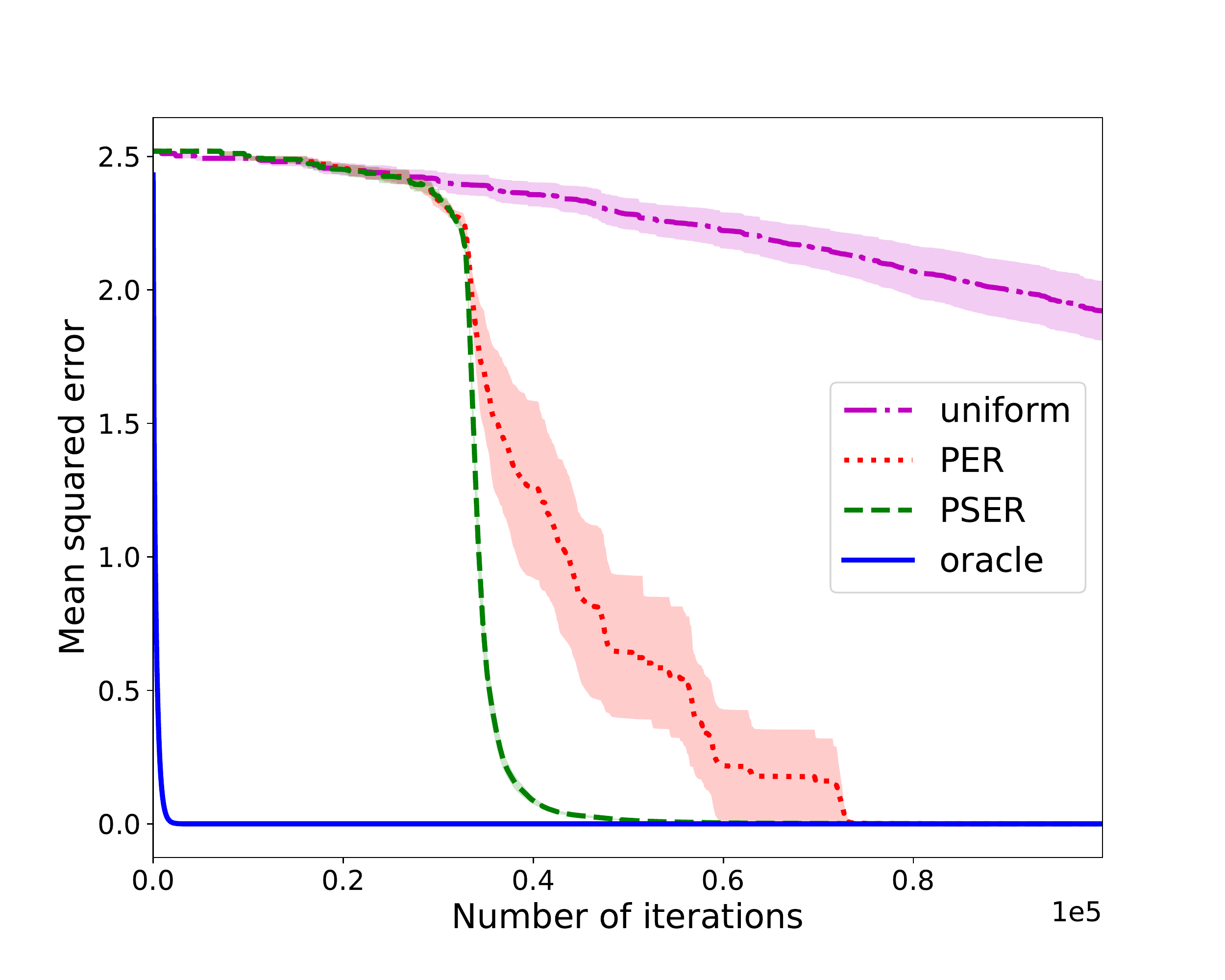}
}%
\hspace{0.08\textwidth} 
\subcaptionbox{
14 states with all transitions initialized with $\epsilon$ priority.
\label{subfig:14zeroappen}
}{
\includegraphics[width=0.7\columnwidth]{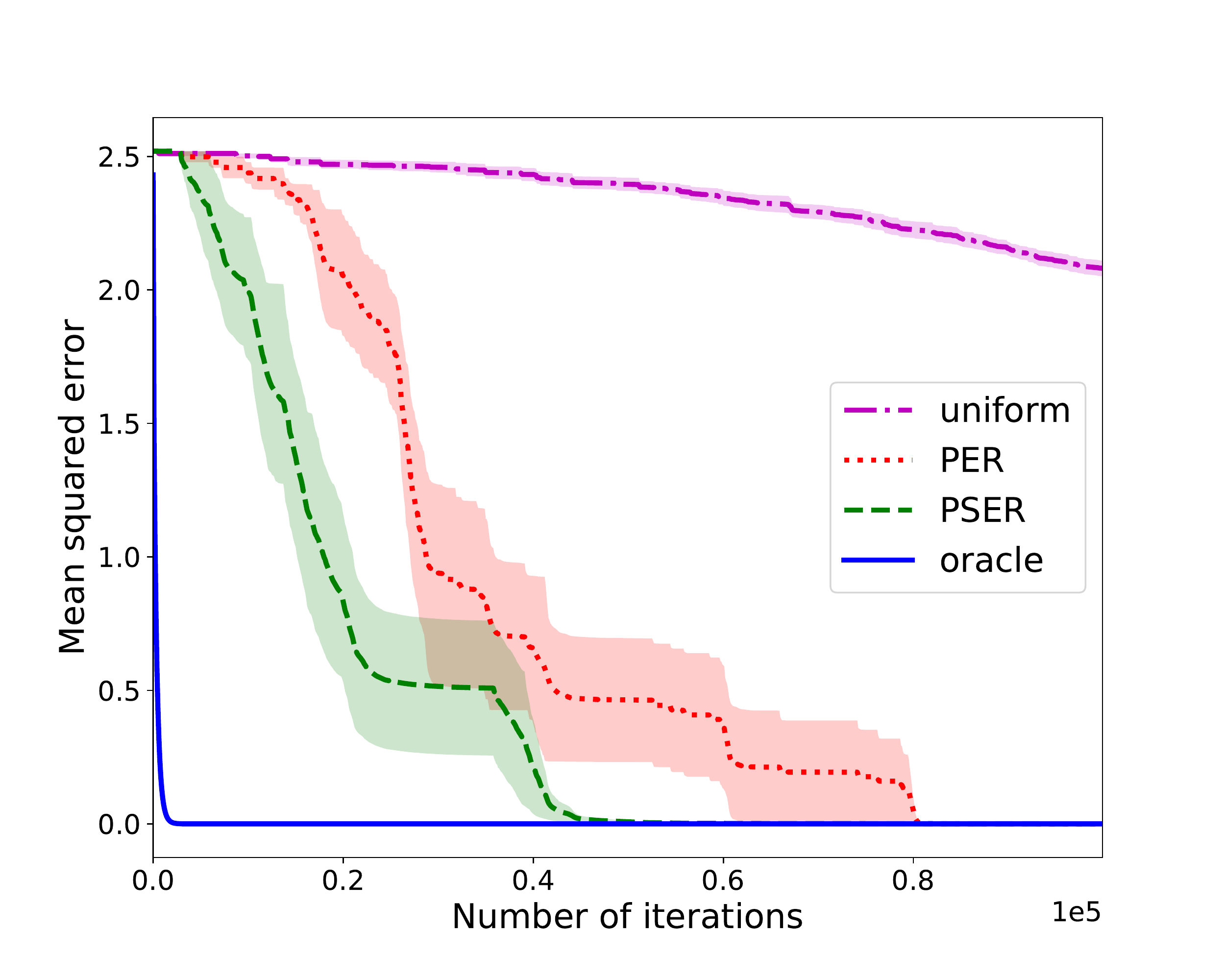}
}%
\hspace{0.08\textwidth} 
\subcaptionbox{
15 states with all transitions initialized with max priority.
\label{subfig:15maxappen}
}{
\includegraphics[width=0.7\columnwidth]{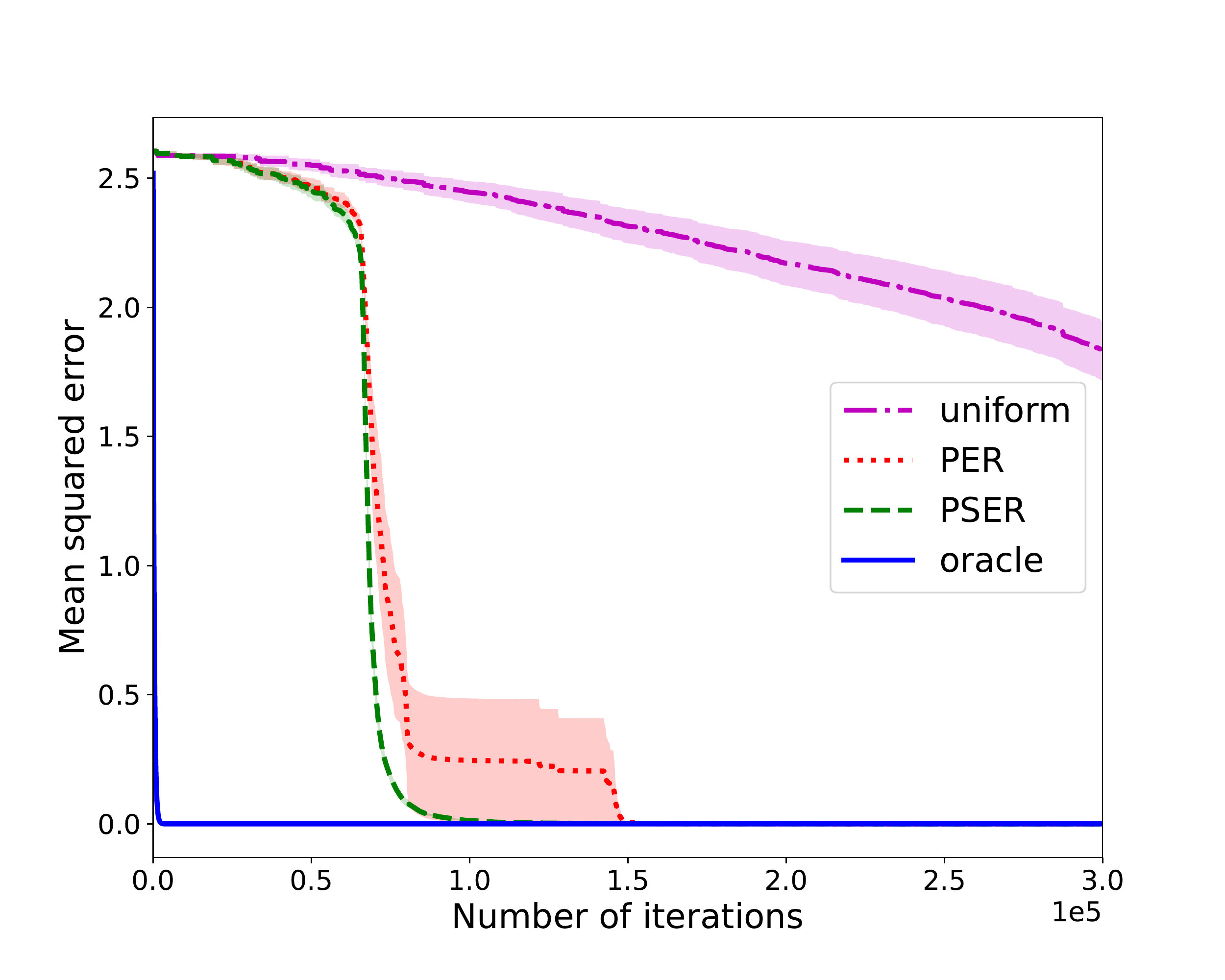}
}%
\hspace{0.08\textwidth} 
\subcaptionbox{
15 states with all transitions initialized with $\epsilon$ priority.
\label{subfig:15zeroappen}
}{
\includegraphics[width=0.7\columnwidth]{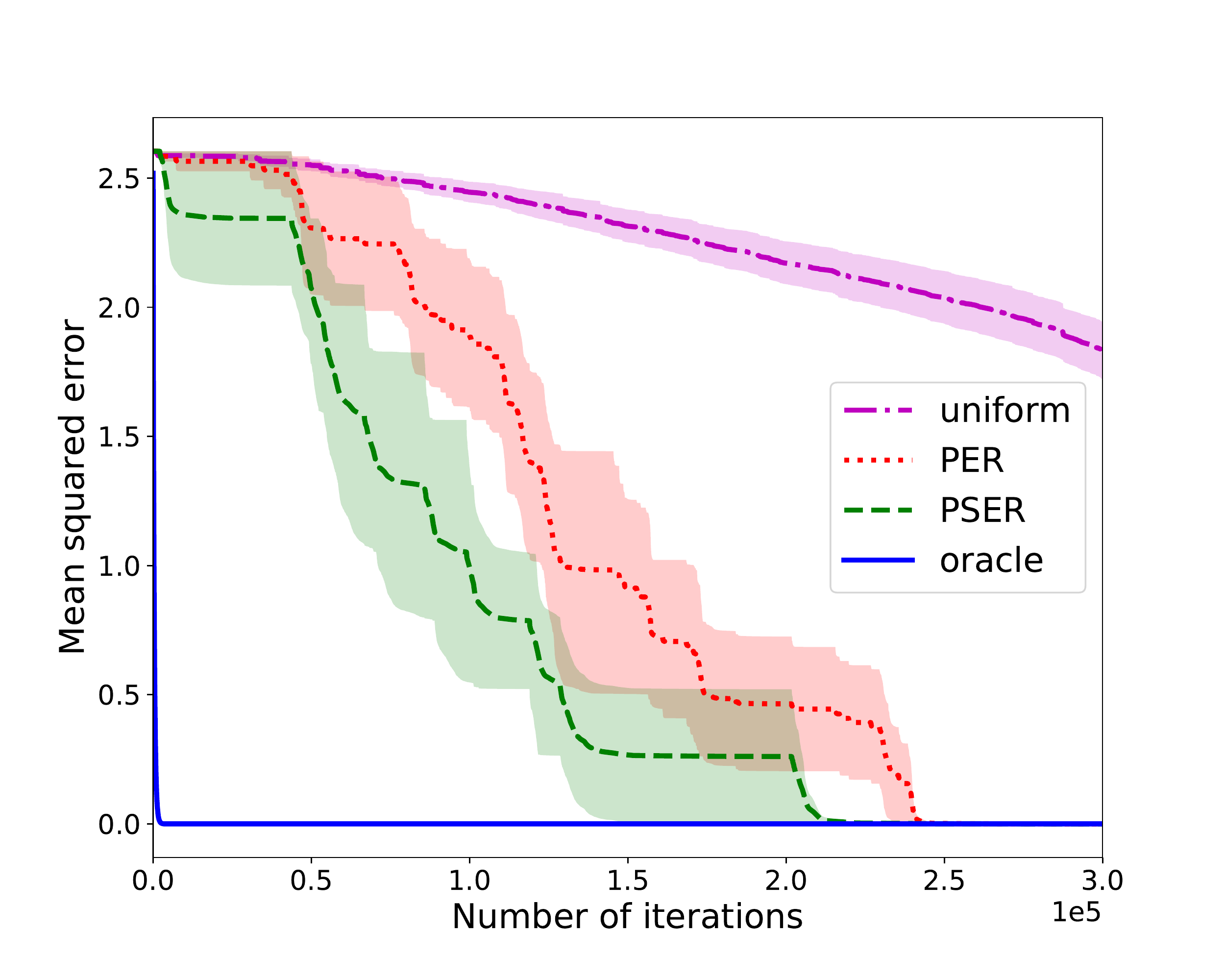}
}%
\hspace{0.08\textwidth} 
\subcaptionbox{
16 states with all transitions initialized with max priority.
\label{subfig:16maxappen}
}{
\includegraphics[width=0.7\columnwidth]{result/17withmax.pdf}
}%
\hspace{0.08\textwidth} 
\subcaptionbox{
16 states with all transitions initialized with $\epsilon$ priority.
\label{subfig:16zeroappen}
}{
\includegraphics[width=0.7\columnwidth]{result/17with0.pdf}
}%
\caption{Results of the Blind Bliffwalk environment comparing the number of iterations until convergence of the true $Q$ value among the PSER, PER, uniform, and oracle agents. We can see that in each case, PSER further improves upon the performance of PER and leads to faster convergence to the true $Q$ value.}
\label{fig:blindcliff_appen}
\end{figure*}

We track the mean squared error (MSE) between the ground truth $Q$ value and $Q$-learning result every 100 iterations. Better performance in this experiment means that the loss curve more closely matches the oracle. The PER paper demonstrated that PER improves performance over uniform sampling. Our results show that PSER further improves the results with much faster and earlier convergence as compared to PER. We show results varying the state-space size from 13 to 16 and and that PSER consistently outperforms PER in this problem as shown in Figure~\ref{fig:blindcliff_appen}.

\section{Evaluation Methodology}

In our Atari experiments, our primary baseline for comparison was Prioritized Experience Replay (PER). For each implementation we used the standard DQN algorithm without any additional modifications to provide a fair comparisons between the different sampling techniques. All of the hyperparameters for DQN were the same between the PSER and PER implementations. The hyperparameters are shown in Table~\ref{DQN_param}.

\subsection{Hyperparameters}
In selecting our final set of hyperparameters for PSER, we tested a range of different values over a subset of Atari games. Table 1 lists the range of values that were tried for each parameter and Table 2 lists the chosen parameters. To obtain the final set of parameters, two parameters were held constant while we tuned one, then we fixed the tuned parameter with best performance and tuned the next. This greatly reduced the search space of parameters and led to a set of parameters that performed well.

\begin{table}[ht]
\caption{PSER hyperparameters tested in experiments.}
\begin{center}
\begin{small}
\begin{sc}
\begin{tabular}{lcccr}
\toprule
Hyperparameter & Range of Values \\
\midrule
 Decay Window $W$    & 5, 10, 20\\
Decay Coefficient $\rho$ & 0.4, 0.65, 0.8\\
Previous Priority $\eta$    & 0, 0.3, 0.5, 0.7\\
\bottomrule
\end{tabular}
\end{sc}
\end{small}
\end{center}
\vskip -0.1in
\end{table}

\begin{table}[h]
\caption{Finalized PSER hyperparameters.}
\label{PSER_param}
\begin{center}
\begin{small}
\begin{sc}
\begin{tabular}{lcccr}
\toprule
Parameter & Value \\
\midrule
Decay Window $W$    & 5\\
Decay Coefficient $\rho$ & 0.4\\
Previous Priority $\eta$    & 0.7\\
\bottomrule
\end{tabular}
\end{sc}
\end{small}
\end{center}
\vskip -0.1in
\end{table}

\begin{table}[H]
\caption{DQN hyperparameters}
\label{DQN_param}
\begin{center}
\begin{small}
\begin{sc}
\begin{tabular}{lcccr}
\toprule
Parameter & Value \\
\midrule
Minibatch size & 32\\
Min history to start learning & 50K frames\\
RMSProp learning rate & 0.00025\\
RMPProp gradient momentum & 0.95\\
Exploration $\epsilon$ & 1.0 $\rightarrow$ 0.01\\
Evaluation $\epsilon$ & 0.001\\
Target Network Period & 10K frames\\
RMSProp $\epsilon$ & 1.0 $\times$ 10$^{-5}$\\
Prioritization type & proportional\\
Prioritization exponent $\alpha$ & 0.5\\
Prioritization I.S. $\beta$ & 0.5 \\
\bottomrule
\end{tabular}
\end{sc}
\end{small}
\end{center}
\vskip -0.1in
\end{table}

\subsection{Normalization}
The normalized score for the Atari 2600 games is calculated as in \citep{schaul2015prioritized}:
\begin{equation}
\label{eq:normalization}
\text{score}_{\text{normalized}} = \frac{\text{score}_{\text{agent}}-\text{score}_{\text{random}}}{|\text{score}_{\text{human}}-\text{score}_{\text{random}}|}.
\end{equation}

We have listed the reported Human and Random scores that were used for normalization in Table~\ref{full_results_table}.

For the ablation study presented in Figure~\ref{ablation_study}, we normalized the results based on the maximum and minimum values achieved during the ablation study, for each game.

\section{Psuedocode}

Algorithm 1 lists the pseudocode for the PSER algorithm. Note this pseudo code closely follows the pseudo code from \cite{schaul2015prioritized} and the difference is how we update the priority for transitions in replay memory.

 \begin{algorithm*}[h]
   \caption{Prioritized Sequence Experience Replay (PSER)}
   \label{alg:algorithm1}
 \begin{algorithmic}
   \STATE {\bfseries Input:} minibatch $k$, step-size $\xi$, replay period $K$ and size $N$, exponents $\alpha$ and $\beta$, budget $T$, decay window $W$, decay coefficient $\rho$, previous priority $\eta$. \\
   Initialize replay memory $\mathcal{H}=\emptyset$, $\Delta=0$, $p_1=1$ \\
   Observe $S_0$ and choose $A_0 \sim\pi_\theta(S_0)$
   \FOR{$t=1$ {\bfseries to} $T$}
   \STATE Observe $S_t, R_t, \gamma_t$
   \STATE Store transition $(S_{t-1}, A_{t-1}, R_{t-1}, \gamma_t, S_t)$ in $\mathcal{H}$ with maximal priority $p_t = \max_{i<t} p_i$
   \IF{$t\equiv 0$  mod $K$}
   \FOR{$j=1$ {\bfseries to} k}
   \STATE Sample transition $j \sim P ( j ) = p _ { j } ^ { \alpha } / \sum _ { i } p _ { i } ^ { \alpha }$
   \STATE Compute importance-sampling weight $w _ { j } = ( N \cdot P ( j ) ) ^ { - \beta } / \max _ { i } w _ { i }$
   \STATE Compute TD-error \\
   $\delta _ { j } = R_{ j - 1 } + \gamma_{ j } \max_a Q _ {\mathrm{target}} ( S_{ j } , a) - Q ( S _ { j - 1 } , A _ { j - 1 } )$
   \STATE Update transition priority $p_j\leftarrow \max \{|\delta_j|+\epsilon, \eta \cdot p_j\}$
   \STATE Accumulate weight-change $\Delta \leftarrow \Delta + w _ { j } \cdot \delta _ { j } \cdot \nabla _ { \theta } Q ( S _ { j - 1 } , A _ { j - 1 } )$
   \FOR{$l=1$ {\bfseries to} $W$}
   \STATE Update transition priority $p_{j-l} \leftarrow \max \{ (|\delta_j| + \epsilon)\cdot\rho^l, p_{j-l} \}$ to transition $l$ steps backward
   \ENDFOR
   \ENDFOR

   \STATE Update weights $\theta \leftarrow \theta + \xi\cdot \Delta$, reset $\Delta = 0$
   \STATE From time to time copy weights into target network $\theta _ { \mathrm { target } } \leftarrow \theta$
   \ENDIF
   \STATE Choose action $A_t \sim \pi_\theta (S_t )$
   \ENDFOR
 \end{algorithmic}
 \end{algorithm*}

\section{Full results}

\begin{figure*}[t]
\centering
\includegraphics[width=.9\textwidth]{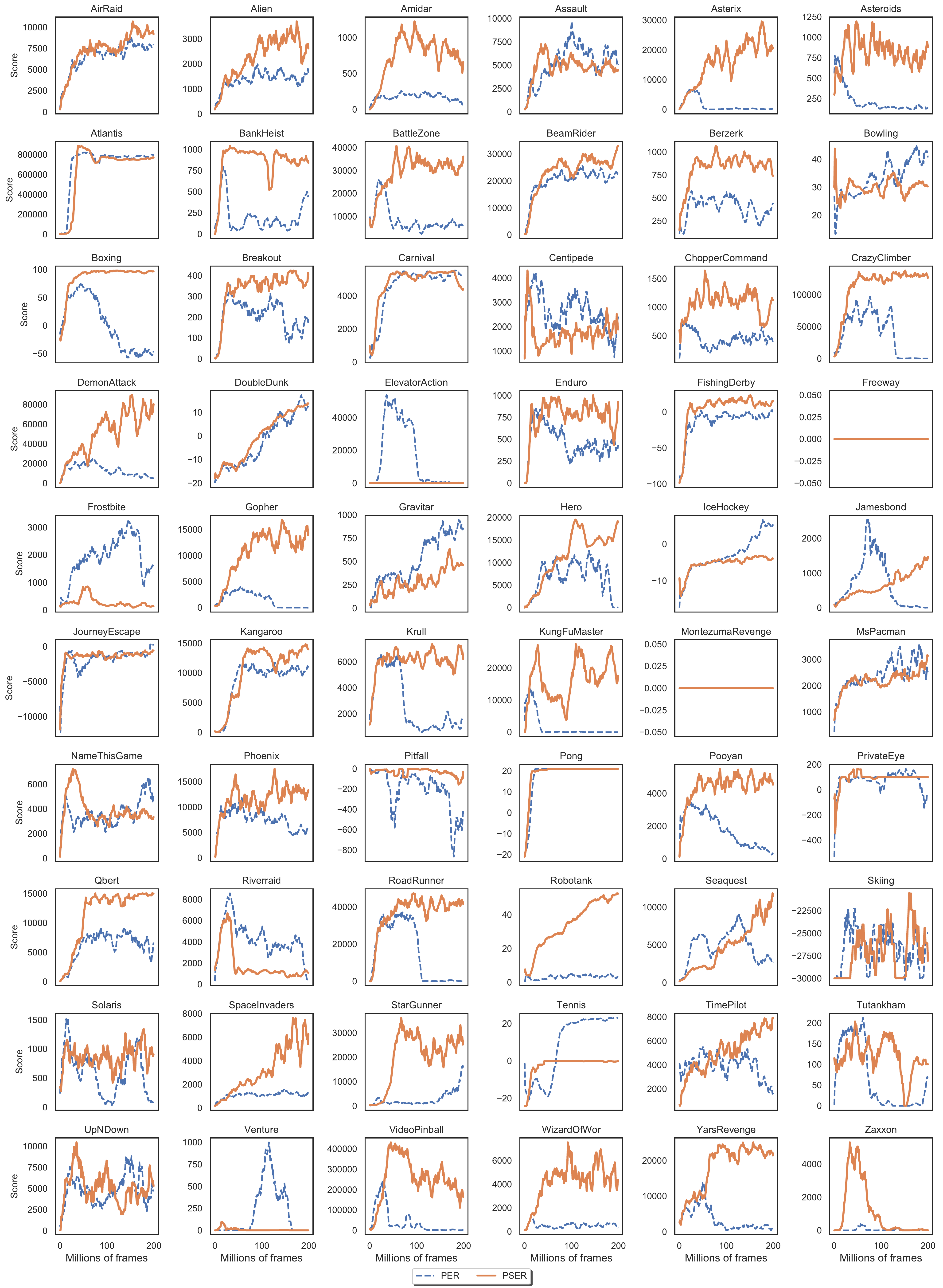}
\caption{Learning curves for DQN with PSER (orange) and DQN with PER (blue) for all 60 games of the Atari 2600 benchmark. Each curve corresponds to a single training run over 200 million unique frames with a moving average smoothed over 10 million frames for clarity.}
\label{full_results}
\end{figure*}

We now present the full results on the Atari benchmark. Figure~\ref{full_results} shows the detailed learning curves of all 60 Atari games using the no-op starts testing regime. These learning curves are smoothed with a moving average of 10M frames to improve readability. In each Atari game, DQN with PER and DQN with PSER are presented.

\begin{figure*}[t]
\centering
\includegraphics[width=.9\textwidth]{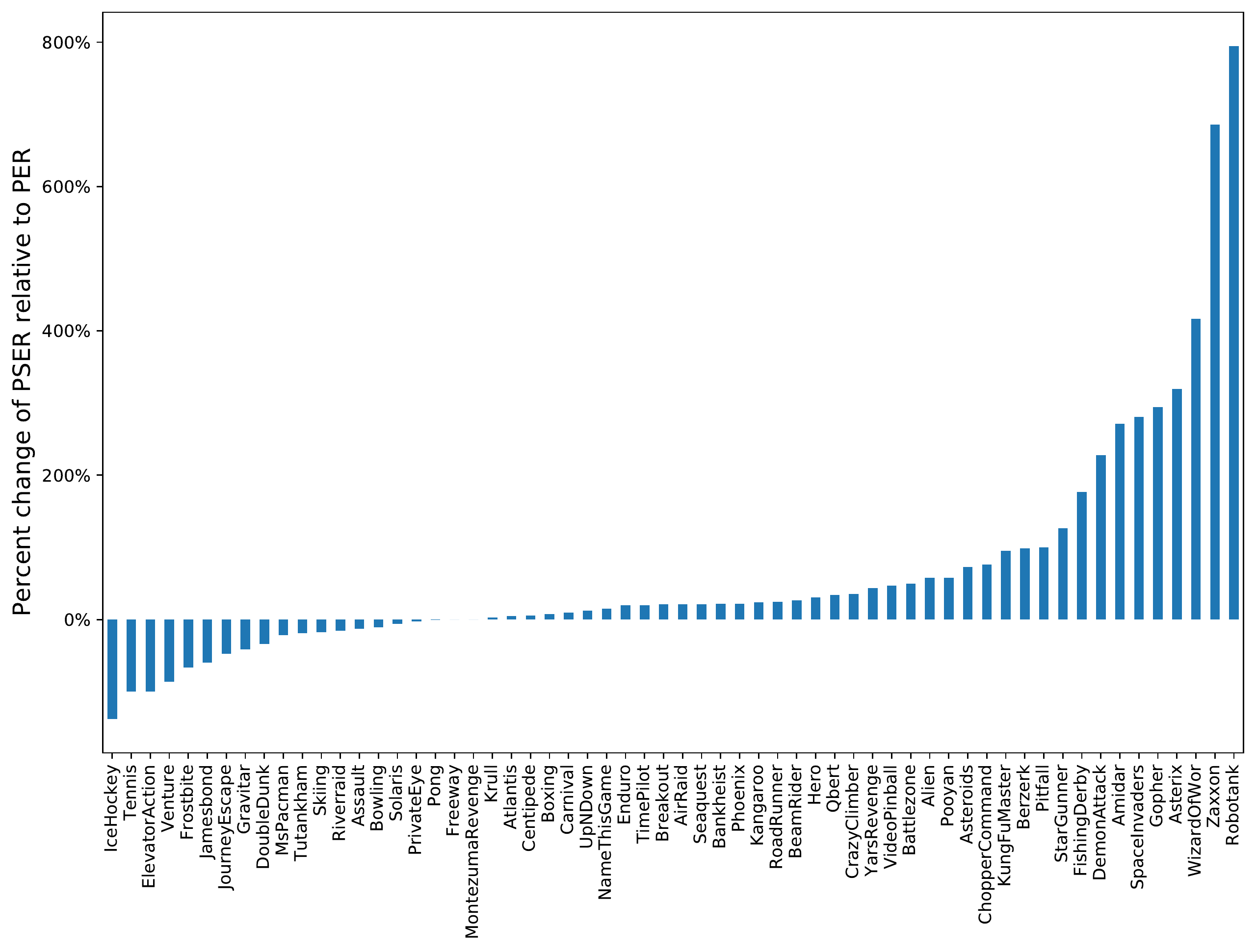}
\caption{Percent change of PSER to PER in all 60 Atari 2600 benchmark games. 0\% on the vertical axis implies equivalent performance; positive numbers represent the cases where PSER performed better; negative numbers represent the cases where PSER performed worse.}
\label{percent_change}
\end{figure*}

Figure~\ref{percent_change} shows the percent change of PSER baselined against PER. Here we present the results from all 60 Atari games, as the scores are not human normalized. The percent change is calculated as follows:
\begin{equation}
\label{eq:change}
\text{Percent Change} = \frac{\text{score}_{\text{PSER}}-\text{score}_{\text{PER}}}{\text{score}_{\text{PER}}}.
\end{equation}
Table~\ref{full_results_table} presents a breakdown of the best scores achieved by each algorithm on all 55 Atari games where human scores were available. Bolded entries within each row highlight the result with the highest performance between PSER and PER.

\begin{table*}[t]
\caption{\textbf{no-op} starts evaluation regime: Here we report the raw scores across all games, averaged over 200 evaluation episodes, from the agent snapshot that obtained the highest score during training. PER and PSER were evaluated using the DQN algorithm described in \citep{mnih2015human}.}
\label{full_results_table}
\vskip 0.15in
\begin{center}
\begin{scriptsize}
\begin{sc}
\begin{tabularx}{0.52\textwidth}{l | c | c | c | c }
\toprule
Game & Random & Human & PER & PSER\\
\midrule
AirRaid & - & - & 8,660.8 & \textbf{10,504.2}\\
Alien & 227.8 & 7,127.7 & 2,724.1 & \textbf{4,297.9}\\
Amidar & 5.8 & 1,719.5 & 364.2 & \textbf{1,351.9}\\
Assault & 222.4 & 742.0 & \textbf{7,761.3} & 6,758.1\\
Asterix & 210.0 & 8,503.3 & 7,806.0 & \textbf{32,766.5}\\
Asteroids & -719.1 & 47,388.7 & 905.6 & \textbf{1,566.8}\\
Atlantis & 12,850.0 & 29,028.1 & 810,043.0 & \textbf{848,064.5}\\
BankHeist & 14.2 & 753.1 & 894.5 & \textbf{1,091.5}\\
BattleZone & 2,360.0 & 37,187.5 & 26,215.0 & \textbf{39,195.0}\\
BeamRider & 363.9 & 16,926.5 & 24,100.2 & \textbf{30,548.9}\\
Berzerk & 123.7 & 2,630.4 & 618.5 & \textbf{1,228.0}\\
Bowling & 23.1 & 160.7 & \textbf{46.7} & 41.7\\
Boxing & 0.1 & 12.1 & 92.8 & \textbf{99.9}\\
Breakout & 1.7 & 30.5 & 355.1 & \textbf{429.1}\\
Carnival & - & - & 5,560.0 & \textbf{6,086.5}\\
Centipede & 2,090.9 & 12,017.0 &  6,192.1 & \textbf{6,542.0}\\
ChopperCommand & 811.0 & 7,387.8 & 746.5 & \textbf{1,317.5}\\
CrazyClimber & 10,780.5 & 35,829.4 & 104,080.5 & \textbf{140,918.0}\\
DemonAttack & 152.1 & 1,971.0 & 22,711.9 & \textbf{74,366.0}\\
DoubleDunk & -18.6 & -16.4 & \textbf{20.6} & 13.7\\
ElevatorAction & - & - & \textbf{47,825.5} & 75.0\\
Enduro & 0.0 & 860.5 & 753.0 & \textbf{901.4}\\
FishingDerby & -91.7 & -38.7 & 13.1 & \textbf{36.3}\\
Freeway & 0.0 & 29.6 & \textbf{0.0} & \textbf{0.0}\\
Frostbite & 65.2 & 4,334.7 & \textbf{3,501.1} & 1,162.2\\
Gopher & 257.6 & 2,412.5 & 4,446.8 & \textbf{17,524.7}\\
Gravitar & 173.0 & 3,351.4 & \textbf{1,569.5} & 918.8\\
Hero & 1,027.0 & 30,826.4 & 15,678.6 & \textbf{20,447.5}\\
IceHockey & -11.2 & 0.9 & \textbf{7.3} & -2.8\\
Jamesbond & 29.0 & 302.8 & \textbf{3,908.8} & 1,572.0\\
JourneyEscape & - & - & \textbf{7,423.0} & 3,898.5\\
Kangaroo & 52.0 & 3,035.0 & 12,150.5 & \textbf{15,051.0}\\
Krull & 1,598.0 & 2,665.5 & 8,189.1 & \textbf{8,436.6}\\
KungFuMaster & 258.5 & 22,736.3 & 14,673.5 & \textbf{28,658.0}\\
MontezumaRevenge & 0.0 & 4,753.3 & \textbf{0.0} & \textbf{0.0}\\
MsPacman & 307.3 & 6,951.6 & \textbf{4,875.3} & 3,834.3\\
NameThisGame & 2,292.3 & 8,049.0 & 6,398.2 & \textbf{7,370.2}\\
Phoenix & 761.4 & 7,242.6 & 12,465.5 & \textbf{15,228.2}\\
Pitfall & -229.4 & 6,463.7 & -8.8 & \textbf{0.0}\\
Pong & -20.7 & 14.6 & \textbf{21.0} & \textbf{21.0}\\
Pooyan & - & - & 3,802.2 & \textbf{6,013.2}\\
PrivateEye & 24.9 & 69,571.3 & \textbf{253.0} & 247.0\\
Qbert & 163.9 & 13,455.0 & 11,463.1 & \textbf{15,396.1}\\
Riverraid & 1,338.5 & 17,118.0 & \textbf{9,684.4} & 8,169.9\\
RoadRunner & 11.5 & 7,845.0 & 41,578.5 & \textbf{51,851.0}\\
Robotank & 2.2 & 11.9 & 5.9 & \textbf{52.7}\\
Seaquest & 68.4 & 42,054.7 & 8,547.4 & \textbf{10,375.2}\\
Skiing & -17,098.1 & -4,336.9 & \textbf{-8,343.3} & -9,807.1\\
Solaris & 1,236.3 & 12,326.7 & \textbf{1,331.7} & 1,253.2\\
SpaceInvaders & 148.0 & 1,668.7 & 1,774.0 & \textbf{6,754.8}\\
StarGunner & 664.0 & 10,250.0 & 15,672.0 & \textbf{35,448.5}\\
Tennis & -23.8 & -8.3 & \textbf{23.7} & 0.0\\
TimePilot & 3,568.0 & 5,229.2 & 7,545.0 & \textbf{9,033.5}\\
Tutankham & 11.4 & 167.6 & \textbf{223.3} & 180.9\\
UpNDown & 533.4 & 11,693.2 & 10,786.2 & \textbf{12,098.9}\\
Venture & 0.0 & 1,187.5 & \textbf{1,115.0} & 152.5\\
VideoPinball & 16,256.9 & 17,667.9 & 232,144.6 & \textbf{340,562.7}\\
WizardOfWor & 563.5 & 4,756.5 & 1,674.0 & \textbf{8,644.5}\\
YarsRevenge & 3,092.9 & 54,576.9 & 19,538.9 & \textbf{28,049.8}\\
Zaxxon & 32.5 & 9,173.3 & 790.0 & \textbf{6,207.5}\\

\bottomrule
\end{tabularx}
\end{sc}
\end{scriptsize}
\end{center}
\vskip -0.1in
\end{table*}

\end{document}